\documentclass[letterpaper]{article} 
\usepackage[preprint]{aaai2027}  
\usepackage[hyphens]{url}  
\usepackage{graphicx} 
\usepackage{natbib}  
\usepackage{caption} 
\usepackage{algorithm}
\usepackage{algorithmic}

\usepackage{newfloat}
\usepackage{listings}
\DeclareCaptionStyle{ruled}{labelfont=normalfont,labelsep=colon,strut=off} 
\floatstyle{ruled}
\newfloat{listing}{tb}{lst}{}
\floatname{listing}{Listing}

\usepackage{booktabs}

\usepackage{amsmath}
\usepackage{amssymb}
\usepackage{bbding} 

\title{ViTaR: Visuo-Tactile Residual Adaptation for Foundation VLA Manipulation}
\author{
    Yi Wang\textsuperscript{1,*},
    Renjun Wu\textsuperscript{1,*},
    Jinyan Liu\textsuperscript{1},
    Xuesong Li\textsuperscript{1,\Envelope}
}
\affiliations{
    \textsuperscript{1}Beijing Institute of Technology\\
    \textsuperscript{*}Equal contribution\quad
    \textsuperscript{\Envelope}Corresponding author\\
    Project page: \url{https://icr-lab.github.io/ViTaR}
}

\begin{document}

\maketitle

\begin{abstract}
As Vision-Language-Action (VLA) models scale toward real-world deployment, contact-rich manipulation exposes a critical blind spot: these policies encode broad visual-semantic priors yet remain unaware of local contact events, producing identical actions whether contact is established, lost, or destabilized. Existing remedies either modify VLA internals, risking catastrophic forgetting, or demand online reinforcement under near-failure contact conditions. Both grant tactile unbounded influence over action generation, conflicting with the priors that make VLAs generalizable. We introduce ViTaR, which reframes tactile feedback from an action-generating perceptual input to an execution modulator that selects and scales bounded residual corrections atop a frozen VLA, preserving pretrained capabilities by construction. ViTaR decomposes adaptation into two stages: Effect-Guided Modeling determines \emph{whether} and \emph{which} correction is locally justified via outcome-grounded preference evidence, and Residual Action Modulation converts this evidence into a residual choice with continuously scaled gain from real-time visuotactile observations. On the UniVTAC benchmark spanning seven contact-rich tasks, ViTaR achieves 61.3\% average success, a 30.6 percentage-point improvement over its frozen VLA base that also surpasses purpose-built tactile baselines. Physical-robot experiments confirm that bounded tactile modulation transfers to real sensor noise and dynamics.
\end{abstract}

\begin{figure*}[t]
    \centering
    \includegraphics[width=\textwidth]{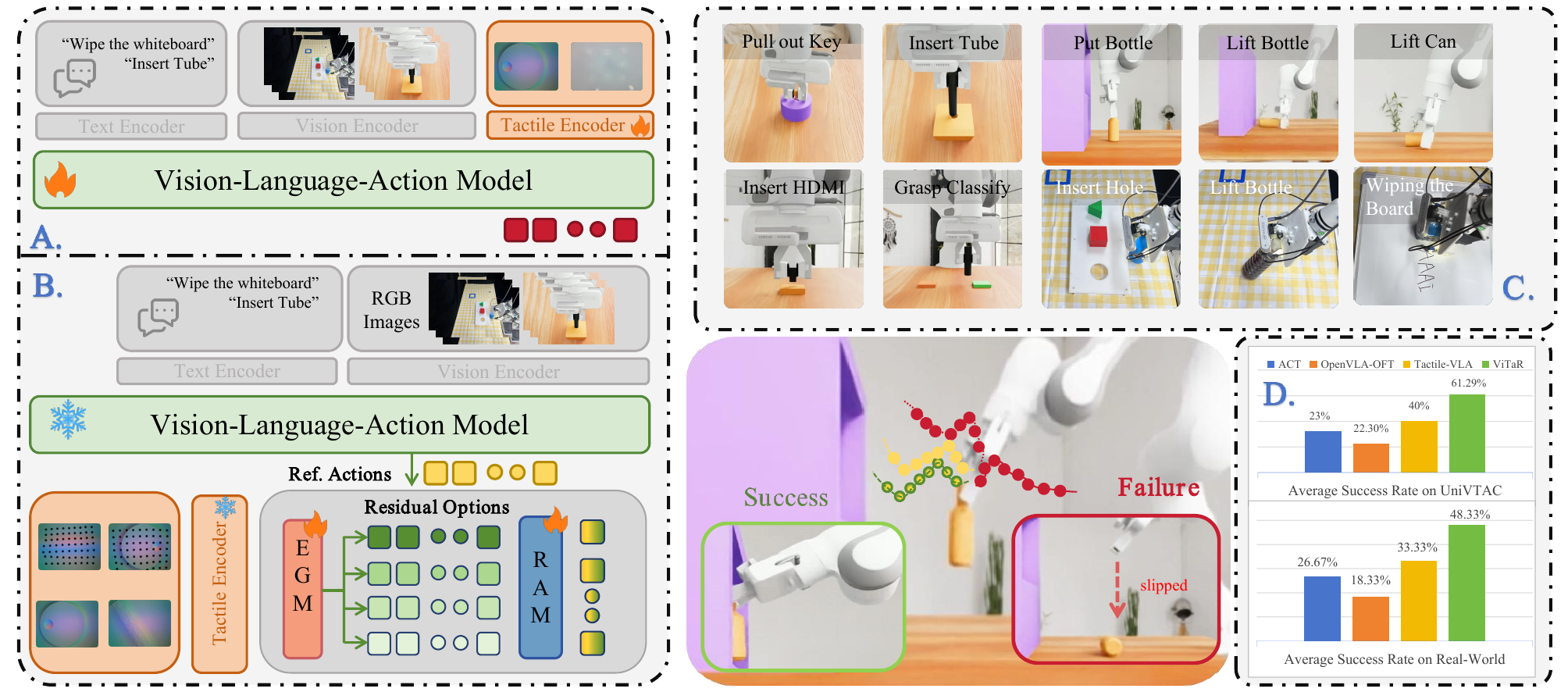}
    \caption{ViTaR at a glance. (A) A paradigm for introducing tactile modality into a pretrained VLA. (B) The tactile-modality integration paradigm proposed by ViTaR. (C) The seven contact-rich UniVTAC tasks and three physical-robot contact-rich tasks; task names shown in white indicate physical-robot tasks. (D) Task success rates of baselines and ViTaR across UniVTAC and physical-robot tasks; grouped bars follow the fixed left-to-right order ACT, OpenVLA-OFT, Tactile-VLA, and ViTaR.}
    \label{fig:teaser}
\end{figure*}

\section{Introduction}

Vision-Language-Action (VLA) models encode broad visual-semantic priors from large-scale pretraining~\cite{brohan2023rt1,brohan2023rt2,kim2024openvla,black2024pi0,chi2023diffusion,physicalintelligence2025pi05,ghosh2024octo,wang2024hpt,vla-adapter}, yet contact-rich tasks such as insertion, connector mating, and force-sensitive grasping depend on resolving local physical interactions that external cameras cannot observe~\cite{cao2026tactilesurvey,huang2025tactilevla}. As these models are deployed to increasingly contact-intensive settings, this gap becomes a critical bottleneck. Once deployed, a frozen VLA is blind to contact dynamics: it commits to the same action regardless of whether contact is established, slipping, or lost.

Existing approaches to incorporating tactile feedback share a common structural commitment: they grant touch \emph{unbounded} influence over action generation. Methods that fuse tactile into VLA internals~\cite{huang2025tactilevla,li2026atvla,morissette2026tacfilm} risk catastrophic forgetting, while reinforcement-based adaptation~\cite{ma2026taccorl,zheng2026torlvla} demands exploratory rollouts at near-failure contact states where safety margins are thinnest. In both paradigms, tactile can override any dimension of the output action, directly conflicting with the visual-semantic priors that make VLAs generalizable.

Our key observation is that a frozen VLA already provides the correct semantic action \emph{direction}; what it lacks is calibration of execution under contact conditions absent during pretraining. This asymmetry motivates a paradigm shift: \textbf{tactile feedback should serve as an execution modulator rather than participating in unconstrained action generation} (Fig.~\ref{fig:teaser}). Realizing this paradigm demands two ingredients: a \emph{correction mechanism} preserving the base action space, and a \emph{learning signal} that ranks corrections by local effect. Residual policies~\cite{johannink2019residual,hoque2021thriftydagger} supply the first, but existing formulations lack contact feedback and assume state-independent corrections. For the learning signal, global reward regression is ill-suited because absolute scores are not comparable across heterogeneous contact states. Within-state preference comparison~\cite{bradley1952rank} resolves this: ranking corrections at the \emph{same} restored contact state recovers relative local effects without cross-state calibration, providing exactly the signal that contact-conditioned residual selection requires.

\begin{figure*}[t]
    \centering
    \includegraphics[width=0.98\textwidth]{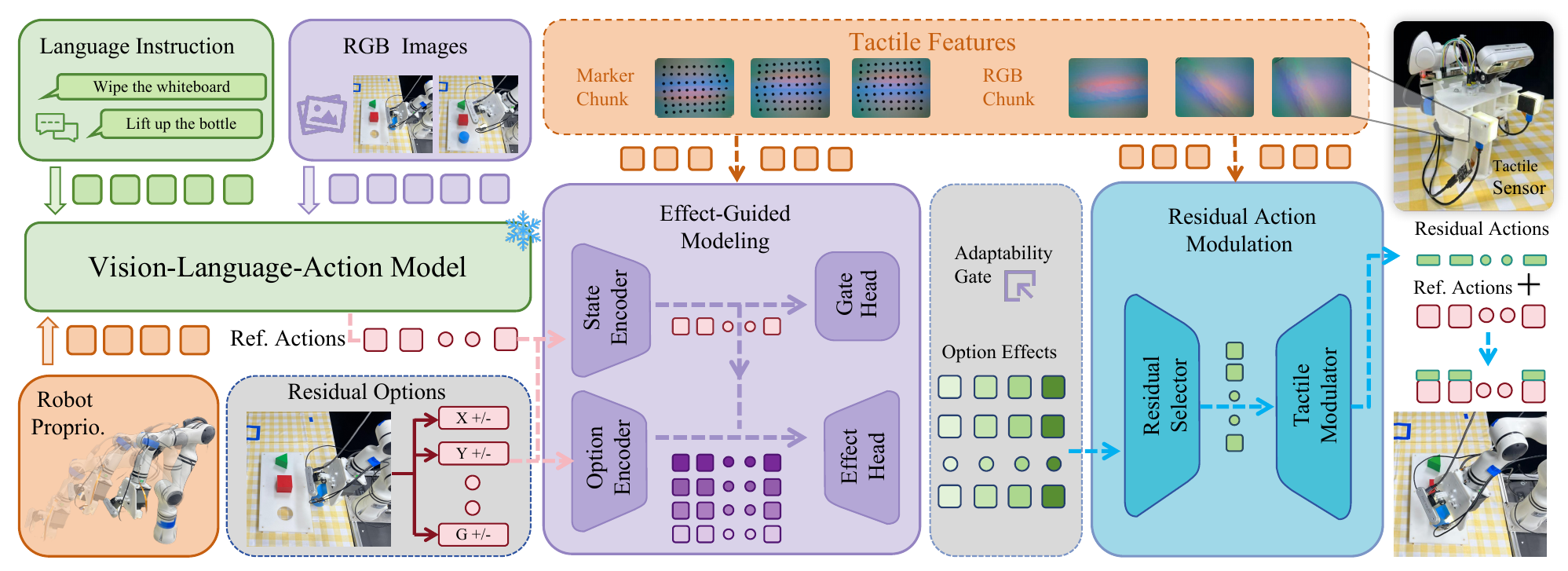}
    \caption{Overview of ViTaR. A frozen VLA maps language, RGB observations, and proprioception to a base action. Marker and tactile-image chunks provide local contact evidence. EGM estimates state adaptability and relative effects for the base action and residual options; RAM then selects an action choice and predicts a continuous scale for the selected residual before execution.}
    \label{fig:vitar_overview}
\end{figure*}

We introduce \textbf{ViTaR} (\textbf{Vi}suo-\textbf{Ta}ctile \textbf{R}esidual Adaptation), which operationalizes this paradigm through two coupled stages (Fig.~\ref{fig:vitar_overview}). \emph{Effect-Guided Modeling} (EGM) assesses whether the current contact state is improvable and ranks available residuals by expected local effect via pairwise outcome comparisons at restored decision points. \emph{Residual Action Modulation} (RAM) converts this evidence into a discrete action choice---retain the base action or select a residual---and predicts a tactile-conditioned gain. The frozen VLA action is preserved whenever contact evidence does not justify intervention.

On the UniVTAC benchmark~\cite{chen2026univtac}, ViTaR achieves 61.3\% average success across seven contact-rich tasks, a 30.6pp improvement over its frozen OpenVLA-OFT~\cite{qiu2023controlling} base, and consistent gains over purpose-built tactile baselines. Physical-robot experiments validate transfer across diverse contact modalities. Our contributions are threefold:
\begin{itemize}
\item We propose ViTaR, a \emph{tactile-as-execution-modulation} framework that augments a frozen VLA with bounded residual corrections selected and scaled by tactile evidence, without generating new action directions or modifying pretrained representations.
\item We introduce Effect-Guided Modeling (EGM), which ranks candidate corrections by observed outcome differences within each contact state via pairwise preference learning, eliminating the need for globally calibrated rewards.
\item We validate ViTaR on the UniVTAC benchmark and physical-robot tasks, demonstrating a 30.6\% gain over the frozen VLA base and successful transfer across insertion, sliding, and grasping tasks in real world.
\end{itemize}

\section{Related Work}

\paragraph{Tactile integration into VLA and visuomotor policies.}
Tactile sensing captures proximal contact evidence that external cameras cannot resolve~\cite{cao2026tactilesurvey}. Recent methods fuse touch into VLA policies through diverse mechanisms: deep token fusion~\cite{huang2025tactilevla}, temporally decoupled injection~\cite{li2026atvla}, feature-level modulation~\cite{morissette2026tacfilm}, and residual tactile representations~\cite{zhang2026restacvla}. Beyond fusion, TacCoRL~\cite{ma2026taccorl} and TORL-VLA~\cite{zheng2026torlvla} use online RL at contact bottlenecks; Tactile-WAM~\cite{wu2026tactilewam} routes tactile within a world-action model; HTD~\cite{niu2026touchdreaming} predicts future tactile states; and TactiDex~\cite{ni2026tactidex} establishes a tactile-guided dexterous benchmark. Foundation tactile encoders~\cite{yuan2026ftp1,higuera2024sparsh} and large-scale visuotactile datasets~\cite{hua2026vtouch} further support these efforts. Despite their diversity, all above methods treat tactile as a perceptual modality participating in action generation with unbounded influence. ViTaR departs from this design: it constrains tactile to a bounded execution-modulation role, selecting and scaling structured residuals atop the frozen base policy without altering its pretrained representations.

\paragraph{Residual and intervention policy learning.}
Residual policy learning augments a fixed base policy with a learned correction term, preserving base competence while enabling task-specific adaptation~\cite{silver2018residual,johannink2019residual}. Intervention learning~\cite{hoque2021thriftydagger,mandlekar2020iwr} and shared-autonomy frameworks~\cite{haldar2023fish} similarly overlay corrective actions under predefined trigger conditions. These methods validate the value of structured corrections but operate without contact feedback and assume correction directions are independent of the local physical state. ViTaR extends this paradigm along two axes: it conditions both correction selection and magnitude on real-time contact evidence, and grounds the selection in outcome-based preference supervision rather than hand-designed triggers.

\paragraph{Preference-based learning for embodied agents.}
Preference-based reward learning acquires relative quality judgments without globally calibrated scalar rewards~\cite{christiano2017deep,bradley1952rank}. Extensions to embodied settings address online feedback~\cite{lee2021pebble}, demonstration-based extrapolation~\cite{brown2020drex}, unified multi-feedback platforms~\cite{cai2025unirlhf}, annotator heterogeneity~\cite{yuan2026prefmoe}, and risk sensitivity~\cite{tung2026riskaware}. However, existing applications uniformly learn global reward models from cross-episode trajectory comparisons. ViTaR operates at a fundamentally different granularity: it learns \emph{within-state} effect rankings from local branch comparisons at restored decision points, recovering which correction is locally superior without requiring a reward function that generalizes across heterogeneous contact configurations.

\section{Method}

\subsection{Problem Formulation}
\label{sec:problem}

We consider contact-rich manipulation augmented by a frozen pretrained Vision-Language-Action (VLA) policy $\pi_{\mathrm{VLA}}$. At each decision step $t$, the frozen VLA maps observation $o_t$ and language instruction $\ell$ to a base action chunk,
\begin{equation}
    A_t^{\mathrm{ref}}=\pi_{\mathrm{VLA}}(o_t,\ell),
    \qquad
    A_t^{\mathrm{exec}}=A_t^{\mathrm{ref}}+\Delta A_t,
    \label{eq:exec}
\end{equation}
where $A_t^{\mathrm{ref}}$ denotes the frozen VLA output and $\Delta A_t$ is a contact-aware residual correction. An \emph{action choice} either retains the base action ($\Delta A_t=0$) or selects a structured 7D delta $d(u)$ from a small task-aligned candidate set $\mathcal{R}$, whose directions are derived from the predominant signed components of frozen VLA action chunks. For an $H$-step action chunk, the same scaled delta is applied uniformly: $\Delta A_t=(\alpha\,d(u),\ldots,\alpha\,d(u))\in\mathbb{R}^{H\times 7}$.

\subsection{Overview}

ViTaR augments the frozen base policy with two learned components. \emph{Effect-Guided Modeling} (EGM) maps the local state and available action choices to an \emph{adaptability score} $g_i\in\mathbb{R}$ and relative \emph{effect scores} $\{e_i(c)\}_{c\in\mathcal{C}_i}$; \emph{Residual Action Modulation} (RAM) uses them to retain the base action or select and continuously scale a residual. Throughout EGM and RAM, $i$ indexes a local decision point where residuals are evaluated, whereas $t$ indexes a control step in the associated $H$-step action chunk. Thus, $\Delta A_i$ denotes the residual selected at decision point $i$, which instantiates $\Delta A_t$ in Eq.~\eqref{eq:exec} at each step of its chunk.

\begin{figure}[!h]
    \centering
    \includegraphics[width=\columnwidth]{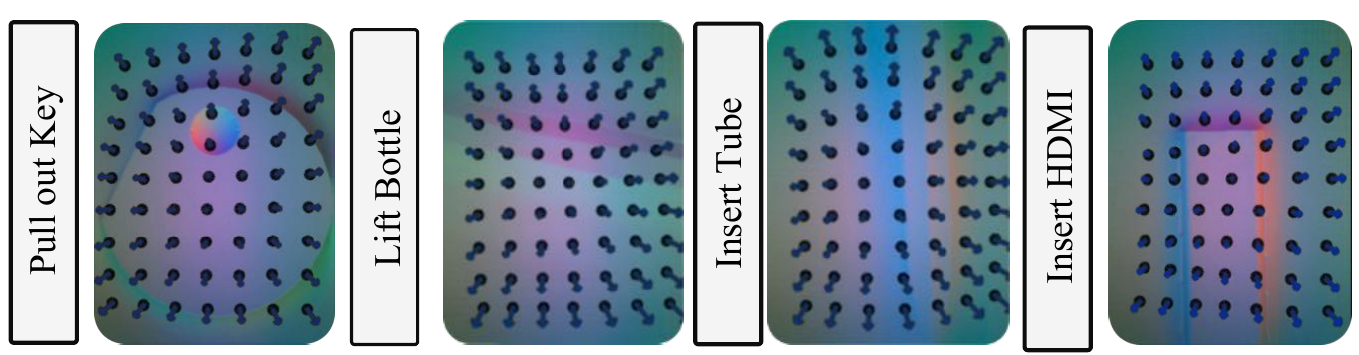}
    \caption{Marker motion on raw simulated GelSight Mini images from UniVTAC. Arrows indicate the displacement directions of markers between pre-contact and post-contact observations. These local motions provide evidence of contact deformation and stability for the marker-derived contact descriptor $m_i$.}
\label{fig:marker_displacement}
\end{figure}

Both modules are supervised by short-horizon branch rollouts restored to common decision points (detailed in \S\ref{sec:egm}). At deployment, only current observations are used: a marker-derived contact descriptor $m_i$ conditions residual selection (Fig.~\ref{fig:marker_displacement}), while a visuotactile summary $z_i$ determines the scaling gain (Fig.~\ref{fig:vitar_overview}).

\begin{figure*}[!t]
    \centering
    \includegraphics[width=0.98\textwidth]{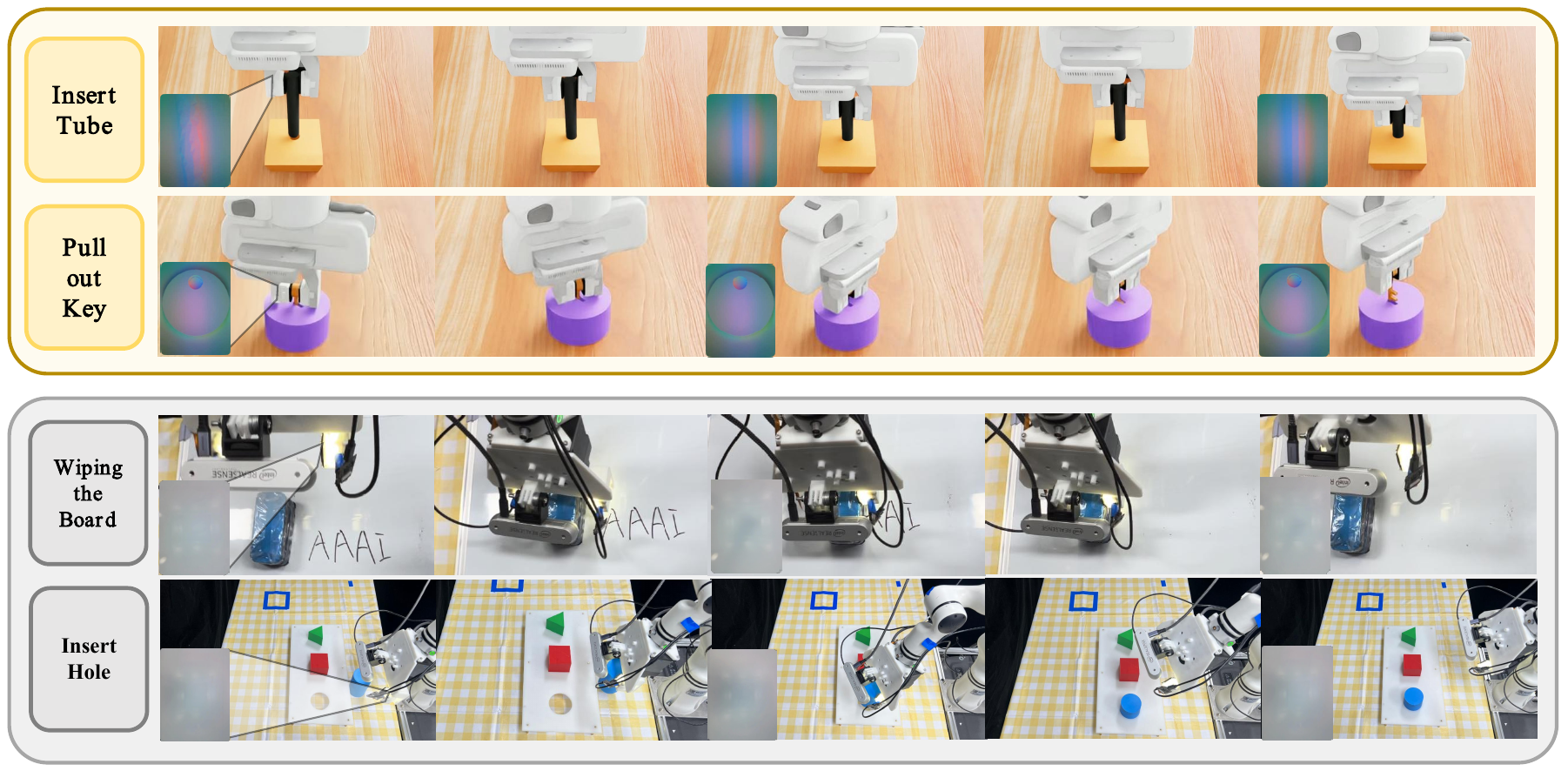}
    \caption{Representative contact-rich task sequences. The upper rows illustrate UniVTAC scenarios spanning insertion, placement, and pulling; the lower rows show the physical \emph{Wiping the Board} and \emph{Insert Hole} tasks. Insets show tactile observations where they are available.}
\label{fig:task_setups}
\end{figure*}

\subsection{Effect-Guided Modeling}
\label{sec:egm}

A residual that improves one contact state may degrade another; EGM therefore learns state-conditioned effect evidence by comparing action choices within the same local state. It serves as a supervision source for RAM rather than a deployed action policy.

\subsubsection{Branch-Outcome Preference Learning}

A \emph{branch} is a short-horizon rollout executed from a restored local state under a single action choice. Let $s_i$ and $x_i$ denote the decision point and its representation, and $c\in\mathcal{C}_i\equiv\{\varnothing\}\cup\mathcal{R}_i$ an action choice. The base choice retains $A_t^{\mathrm{ref}}$; $u\in\mathcal{R}_i$ applies its fixed $d(u)\in\mathbb{R}^{7}$ at every action step. Approximate restoration is handled via reliability weighting; see Eq.~\eqref{eq:preference_weight}.

For each branch, we form a local outcome score from task and contact measurements,
\begin{equation}
    q_i(c)=\mathbf{w}_q^{\top}\boldsymbol{\phi}_i(c),
\label{eq:branch_score}
\end{equation}
where $\boldsymbol{\phi}_i(c)$ contains task outcome, marker-based contact retention, and restoration quality with fixed weights $\mathbf{w}_q$. Because $q_i(c)$ is meaningful only within a single decision group, we compare each residual with the base action at the same decision point,
\begin{equation}
    \Delta q_i(u)=q_i(u)-q_i(\varnothing).
    \label{eq:relative_outcome}
\end{equation}

Using a threshold $\delta=0.05$, we partition the residual options into positive, neutral, and negative sets,
\begin{equation}
\begin{aligned}
\mathcal{U}_i^+ &= \{u\in\mathcal{R}_i:\Delta q_i(u)>\delta\},\\
\mathcal{U}_i^0 &= \{u\in\mathcal{R}_i:|\Delta q_i(u)|\le\delta\},\\
\mathcal{U}_i^- &= \{u\in\mathcal{R}_i:\Delta q_i(u)<-\delta\}.
\end{aligned}
\label{eq:preference_sets}
\end{equation}
This induces preference relations: positive $\succ$ base $\succ$ negative, and positive $\succ$ neutral; neutral options are not directly compared with the base action.

We weight each preference by both its score gap and the quality of its restored branches,
\begin{equation}
\begin{aligned}
w_{i,c_i^+,c_i^-}
&=\max\!\{\delta,q_i(c_i^+)-q_i(c_i^-)\}
  \kappa_i(c_i^+)\kappa_i(c_i^-),\\
\kappa_i(c)
&=\begin{cases}
1, & c=\varnothing\ \text{or restoration checks pass},\\
0.25, & \text{otherwise}.
\end{cases}
\end{aligned}
\label{eq:preference_weight}
\end{equation}
This weighting favors large outcome gaps from reliably restored branches without treating approximate restoration as exact counterfactual evidence. We fit the scalar comparator $e_i(c)$ with a weighted Bradley--Terry objective \cite{bradley1952rank},
\begin{equation}
    \mathcal{L}_{\mathrm{rank}}
    =
    \mathbb{E}_{i}\,\mathbb{E}_{(c_i^+,c_i^-)}
    \left[-w_{i,c_i^+,c_i^-}\log\sigma\!\left(e_i(c_i^+)-e_i(c_i^-)\right)\right].
    \label{eq:rank_loss}
\end{equation}
The resulting objective learns local effect orderings without requiring a globally consistent branch-score scale; the post-rollout scores $q_i(c)$ serve only as offline supervision targets.

\subsubsection{Adaptability and Effect Scoring}

The state representation $x_i$ concatenates the marker-derived contact descriptor $m_i$, visual-proprioceptive observations, language embedding, execution history, and a base-action encoding $\phi_A(A_i^{\mathrm{ref}})$. Crucially, all components of $x_i$ are available at inference time, unlike the post-rollout terms in $\boldsymbol{\phi}_i(c)$ that are used only for offline target construction. EGM predicts
\begin{equation}
    g_i=G_\theta(x_i),
    \qquad
    e_i(c)=E_\theta(x_i,c).
    \label{eq:egm_scores}
\end{equation}
$g_i$ estimates the potential for a useful residual; $e_i(u)-e_i(\varnothing)$ compares a residual with the base action under the current contact state.

The positive set in Eq.~\eqref{eq:preference_sets} directly defines the binary adaptability target,
\begin{equation}
y_i^{\mathrm{adapt}}=
\begin{cases}
1, & \mathcal{U}_i^+\neq\varnothing,\\
0, & \text{otherwise}.
\end{cases}
\label{eq:adaptability_target}
\end{equation}
We optimize the per-decision binary cross-entropy
\[
\mathcal{L}_{\mathrm{adapt}}
=
-y_i^{\mathrm{adapt}}\log\sigma(g_i)
-(1-y_i^{\mathrm{adapt}})\log\!\left(1-\sigma(g_i)\right),
\]
where $\sigma$ is the logistic sigmoid. The EGM objective is
\begin{equation}
    \mathcal{L}_{\mathrm{EGM}}
    =
    \mathcal{L}_{\mathrm{rank}}
    +\lambda_{\mathrm{adapt}}\mathcal{L}_{\mathrm{adapt}}.
    \label{eq:egm_obj}
\end{equation}
$\lambda_{\mathrm{adapt}}$ balances the two objectives. EGM is trained to convergence and frozen before its outputs are used to construct RAM supervision targets.

\begin{table*}[!t]
\centering
\begin{tabular*}{\textwidth}{@{\extracolsep{\fill}}lcccccccc@{}}
\toprule
Method & \shortstack{Lift\\Bottle} & \shortstack{Pull-out\\Key} & \shortstack{Lift\\Can} & \shortstack{Put\\Bottle} & \shortstack{Insert\\HDMI} & \shortstack{Insert\\Tube} & \shortstack{Grasp\\Classify} & Avg. \\
\midrule
ACT                & 42.0\% & 28.0\% & 20.0\% & 28.0\% & 15.0\% & 45.0\% & 50.0\% & 32.6\% \\
VITaL              & 72.0\% & 47.0\% & 8.0\% & 32.0\% & 6.0\% & 34.0\% & \textbf{100.0\%} & 42.7\% \\
ACT + UniVTAC      & 71.0\% & 46.0\% & 29.0\% & 31.0\% & 28.0\% & 56.0\% & 99.0\% & 51.4\% \\
Tactile-VLA        & \textbf{97.0\%} & 32.0\% & 15.0\% & 10.0\% & 12.0\% & 56.0\% & 58.0\% & 40.0\% \\
$\pi_{0.5}$       & 92.0\% & 38.0\% & \textbf{70.0\%} & 14.0\% & 18.0\% & 38.0\% & 68.0\% & 48.3\% \\
OpenVLA-OFT (ref.) & 45.0\% & 33.0\% & 21.0\% & 8.0\% & 9.0\% & 55.0\% & 44.0\% & 30.7\% \\
\midrule
ViTaR (ours)       & 88.0\% & \textbf{55.0\%} & 44.0\% & \textbf{35.0\%} & \textbf{38.0\%} & \textbf{69.0\%} & \textbf{100.0\%} & \textbf{61.3\%} \\
\bottomrule
\end{tabular*}
\caption{Success rate (\%) on seven contact-rich UniVTAC tasks. ViTaR operates on top of the frozen OpenVLA-OFT base policy. Best reported result in each column is in bold.}
\label{tab:main_results}
\end{table*}

\subsection{Residual Action Modulation}
\label{sec:ram}

RAM converts EGM's within-state preference evidence into two deployment-time decisions: which action choice to execute, and at what tactile-conditioned gain to apply the selected residual.

\subsubsection{Preference-Guided Residual Selection}

Because EGM is trained to rank choices within a single branch group, its raw scores and score gaps are ordinal within each state and should not be compared across states. Let $r_i^{\mathrm{EGM}}(c)\in[0,1]$ denote the normalized ordinal rank of $c$ in the ordering induced by $\{e_i(c):c\in\mathcal C_i\}$, and let $\mathbf r_i^{\mathrm{EGM}}$ collect these ranks. This encoding preserves only the within-state order of the current option set.

RAM action-choice labels are constructed only within the corresponding branch group:
\begin{equation}
c_i^{\mathrm{tar}}=
\begin{cases}
    \varnothing, & \mathcal{U}_i^+=\varnothing,\\
    \displaystyle\arg\max_{u\in\mathcal{U}_i^+}\Delta q_i(u), & \text{otherwise}.
\end{cases}
\label{eq:selector_target}
\end{equation}
Thus, the target identifies whether to retain the base action or, when a positive residual exists, which observed residual performed best, without comparing outcome margins across states. The selector maps the current state, option set, and EGM evidence to
\begin{equation}
    p_i(a)=\Pi_\psi\!\left(a\mid x_i,\mathcal{R}_i,g_i,
        \mathbf r_i^{\mathrm{EGM}}\right),
    \qquad a\in\{\varnothing\}\cup\mathcal{R}_i.
    \label{eq:ram_select}
\end{equation}
We train $\Pi_\psi$ via cross-entropy against $c_i^{\mathrm{tar}}$; notably, the base action is treated as an ordinary class label rather than a post-hoc fallback gate. At deployment, $u_i^*=\arg\max_a p_i(a)$ uses only current observations and EGM evidence.

\subsubsection{Tactile-Conditioned Residual Scaling}

Given the selected residual direction $u_i^*$, a scaling network predicts the execution gain from $x_i$ and the bilateral tactile summary $z_i$:
\begin{equation}
\begin{aligned}
    \alpha_i=\sigma\!\left(M_\eta(x_i,u_i^*,z_i)\right),
    \\
    \Delta A_i&=
    \begin{cases}
    0, & u_i^*=\varnothing,\\
    \alpha_i d(u_i^*), & \text{otherwise}.
    \end{cases}
\end{aligned}
    \label{eq:ram_delta}
\end{equation}
$z_i$ encodes bilateral tactile images via per-sensor spatial statistics and temporal differences. By design, $z_i$ modulates only the gain $\alpha_i$; it cannot alter the base action, residual direction, or candidate set.

Because branch data is collected at a single residual amplitude, there is no empirical ground truth for the optimal scale. We therefore repurpose the ordinal rank evidence as an auxiliary gain target, acknowledging that it serves as a within-state proxy rather than a physical magnitude:
\begin{equation}
    \alpha_i^{\mathrm{tar}}
    =
    \begin{cases}
    0, & c_i^{\mathrm{tar}}=\varnothing,\\
    r_i^{\mathrm{EGM}}(c_i^{\mathrm{tar}}), & \text{otherwise}.
    \end{cases}
    \label{eq:ram_magnitude_target}
\end{equation}
The combined RAM objective is $\mathcal{L}_{\mathrm{RAM}}=\mathcal{L}_{\mathrm{selector}}+\lambda_\alpha\mathcal{L}_{\mathrm{scale}}$, where $\mathcal{L}_{\mathrm{selector}}$ is cross-entropy for action choice and $\mathcal{L}_{\mathrm{scale}}$ is weighted binary cross-entropy for the auxiliary gain target. Supplementary multiscale rollouts directly evaluate whether the deployed $\alpha_i$ produces favorable outcomes from an evaluated scale set on held-out states.

\section{Experiments}

Having formulated tactile feedback as evidence for selecting and scaling residual corrections around a frozen VLA policy, we now evaluate whether this formulation translates into consistent task-level gains and separately examine the quality of its learned components. The experiments address three questions:

\noindent (1) Does ViTaR improve task success over its frozen VLA base policy across simulated and physical contact-rich tasks?

\noindent (2) Does EGM's within-state preference learning yield reliable effect evidence for selecting residual corrections?

\noindent (3) Do RAM's effect-guided residual selection and tactile-conditioned scaling yield better residual control than direct residual RL?

\subsection{Experimental Setup}
\subsubsection{Benchmarks and Tasks}
UniVTAC~\cite{chen2026univtac} provides synchronized visual, proprioceptive, and tactile observations for seven contact-rich tasks: \emph{Lift Bottle}, \emph{Pull-out Key}, \emph{Lift Can}, \emph{Put Bottle}, \emph{Insert HDMI}, \emph{Insert Tube}, and \emph{Grasp Classify}. We additionally evaluate \emph{Insert Hole}, \emph{Lift Bottle}, and \emph{Wiping the Board} on a physical robot. The first two mirror their UniVTAC counterparts in contact requirements, while \emph{Wiping the Board} introduces sustained sliding contact not present in the simulated suite. Representative sequences appear in Fig.~\ref{fig:task_setups}.

\subsubsection{Hardware Setup}
Physical demonstrations are collected through ExUMI, a VR-operated UMI-style interface~\cite{chi2024umi}, with 9DTact vision-based tactile sensors~\cite{lin2024_9dtact}. The physical system comprises a 6-DoF RealMan RM-65B, a custom tactile gripper driven by a DH Robotics PGIA-series motor, and wrist and third-person RealSense D455 cameras. Supplementary material details the platform and sensing configuration.

\subsubsection{Baselines}
For ACT~\cite{zhao2023aloha}, VITaL~\cite{george2024vital}, and ACT+UniVTAC, we report the task-level results provided by the UniVTAC benchmark~\cite{chen2026univtac}. For Tactile-VLA, we adapt its tactile-fusion architecture~\cite{huang2025tactilevla} to the UniVTAC observation and control interface, testing direct tactile fusion as a baseline. We also evaluate $\pi_{0.5}$~\cite{physicalintelligence2025pi05}, a strong open-source VLA model without tactile input, under the same seven-task protocol. OpenVLA-OFT~\cite{qiu2023controlling} is our frozen base-policy reference, fine-tuned with OFT on 50 trajectories per task. We do not reproduce AT-VLA, TacFiLM, ResTacVLA, TacCoRL, or TORL-VLA~\cite{li2026atvla,morissette2026tacfilm,zhang2026restacvla,ma2026taccorl,zheng2026torlvla}: at the time of evaluation, we could not obtain public implementations and task-specific checkpoints that were reproducible under the shared UniVTAC control interface and physical-robot protocol. We therefore restrict end-to-end comparison to accessible baselines for which a common evaluation interface can be established; the recent methods above are discussed in Related Work rather than treated as directly comparable reproduced systems.

\subsubsection{Evaluation Protocol and Metrics}
We evaluate each UniVTAC method--task pair over 100 episodes and each physical task over 20 trials, constrained by physical execution cost, using end-to-end binary success as the primary metric. The physical protocol tests consistency across insertion, sustained sliding, and grasping rather than precisely estimating any single-task rate. ViTaR uses 20 restored decision groups per task for short-horizon branch supervision, whose aggregate execution horizon is shorter than that of the 50 expert demonstrations used to fine-tune OpenVLA-OFT. For ablations, Rec., Pres., and Unsafe. denote paired recovery, paired preservation, and safety/early termination, respectively; P$>$B and B$>$N are held-out positive-over-base and base-over-negative ordering accuracies; FPR is measured at 80\% recall; and Off.\ T1 is held-out agreement with the best observed positive residual. Supplementary multiscale and option-set controls report scale-bin accuracy, local outcome regret, the fraction of groups with a positive option, and mean best branch gain. All non-success metrics are post-hoc diagnostics and never inputs to the deployed policy.

\begin{figure}[!t]
    \centering
    \includegraphics[width=\columnwidth]{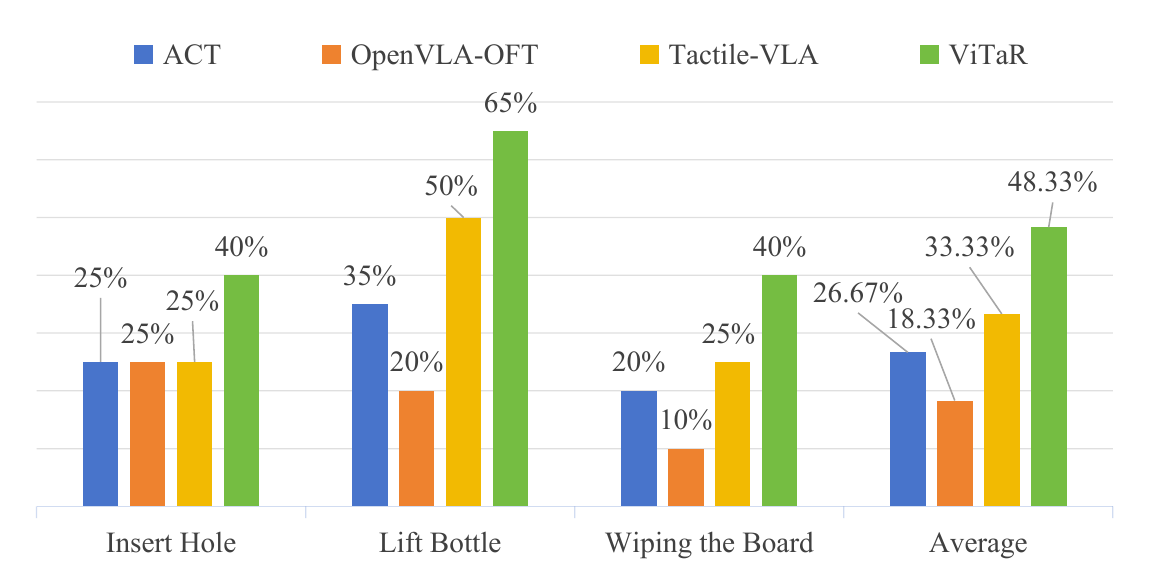}
    \caption{End-to-end success rates on physical-robot contact-rich tasks. Within each group, bars follow the fixed left-to-right order ACT, OpenVLA-OFT, Tactile-VLA, and ViTaR; the rightmost group reports average success.}
    \label{fig:physical_results}
\end{figure}

\subsection{Main Results}

\subsubsection{UniVTAC Success Rates}
Table~\ref{tab:main_results} compares ViTaR with its OpenVLA-OFT base policy and representative imitation, tactile, and foundation-policy baselines on seven contact-rich UniVTAC tasks. ViTaR attains the highest average success rate (61.3\%), doubling the frozen base policy's performance (30.7\% $\to$ 61.3\%, a 30.6pp gain). Because ViTaR retains the base policy's semantic action direction rather than replacing it, its achievable absolute performance remains constrained by the frozen base's semantic action quality. This improvement is achieved without modifying the VLA's pretrained representations, indicating that contact-conditioned residual corrections can substantially improve a frozen policy while preserving its visual-language action prior intact.

Relative to OpenVLA-OFT, ViTaR improves success on all seven UniVTAC tasks (Table~\ref{tab:main_results}), with particularly large gains on Lift Bottle, Put Bottle, Insert HDMI, and Grasp Classify. ViTaR does not dominate every task individually: $\pi_{0.5}$ is stronger on Lift Can (70\% vs.\ 44\%), and Tactile-VLA achieves the highest Lift Bottle result (97\% vs.\ 88\%). We therefore interpret the comparison as broad improvement around the frozen base policy rather than uniform task-level dominance.

\subsubsection{Physical-Robot Success Rates}

Figure~\ref{fig:physical_results} reports end-to-end success on the three physical contact-rich tasks. ViTaR achieves a 48.3\% average success rate, improving on OpenVLA-OFT by 30.0 percentage points, on Tactile-VLA by 15.0 percentage points, and on ACT by 21.7 percentage points. The improvement appears in all three tasks: ViTaR reaches 40.0\% on \emph{Insert Hole}, 40.0\% on \emph{Wiping the Board}, and 65.0\% on \emph{Lift Bottle}. These tasks require qualitatively different contact behaviors (precision alignment, sustained sliding, and stable grasping), so the physical gains reflect the generality of bounded tactile modulation across contact modalities rather than a single task-specific effect. The absolute success level (48.3\%) is bounded by the base policy's semantic action quality; improvements over the base are consistent with the simulated margins. Given the three-task, 20-trial physical protocol, these results provide limited transfer evidence rather than resolving the sim-to-real gap or establishing broad real-world generalization.

\subsection{Ablation Studies}
\label{sec:ablation}

\paragraph{Comparison with direct residual RL.}
Table~\ref{tab:rl_comparison} compares learning formulations around the same frozen OpenVLA-OFT reference, residual coordinates, decision points, and safety limits. PPO-option selects from the same discrete base-action/residual interface, whereas SAC-residual learns a bounded continuous residual without EGM or RAM supervision. ViTaR reaches 61.3\% average success, versus 36.0\% and 32.0\%, and has lower unsafe/early-termination rates (0.04 versus 0.25 and 0.31). Within the shared bounded-residual protocol, these results favor effect-guided residual selection over the tested direct RL formulations. We emphasize that this is a comparison of learning formulations under matched conditions, not a general claim about RL for robotic manipulation.

\begin{table}[!ht]
\centering
\begin{tabular*}{\columnwidth}{@{\extracolsep{\fill}}lcccc@{}}
\toprule
Method & Rec. $\uparrow$ & Pres. $\uparrow$ & Unsafe. $\downarrow$ & Avg. (\%) $\uparrow$ \\
\midrule
PPO-option & .55 & .66 & .25 & 36.0\% \\
SAC-residual & .45 & .74 & .31 & 32.0\% \\
\textbf{Full ViTaR} & \textbf{.79} & \textbf{.92} & \textbf{.04} & \textbf{61.3\%} \\
\bottomrule
\end{tabular*}
\caption{Comparison with direct residual RL. Recovery and preservation are post-hoc matched-start measures; Unsafe. is the safety/early-termination rate.}
\label{tab:rl_comparison}
\end{table}

\begin{table}[!h]
\centering
\begin{tabular*}{\columnwidth}{@{\extracolsep{\fill}}lcccc@{}}
\toprule
Variant & Rec. $\uparrow$ & Unsafe. $\downarrow$ & Avg. $\uparrow$ \\
\midrule
w/o State         & .46 & .16 & 41.0\% \\
w/o Tac.          & .61 & .13 & 46.0\% \\
w/o Both          & .40 & .19 & 36.0\% \\
Fixed Scale        & .57 & .11 & 45.0\% \\
\textbf{Full ViTaR} & \textbf{.79} & \textbf{.04} & \textbf{61.3\%} \\
\bottomrule
\end{tabular*}
\caption{Contact information and continuous scaling ablation. ``w/o State'' removes the marker-derived contact descriptor, ``w/o Tac.'' removes the tactile summary, and ``w/o Both'' removes both. Rec. is paired recovery under matched-start evaluation.}
\label{tab:contact_magnitude}
\end{table}

\paragraph{Contact information and continuous scaling.}
Table~\ref{tab:contact_magnitude} separates the marker-derived contact descriptor $m_i$, used for residual selection, from the tactile summary $z_i$, used for continuous scaling. Fixing the scale reduces success from 61.3\% to 45.0\%. Removing $m_i$ lowers paired recovery from 0.79 to 0.46, while removing $z_i$ lowers success to 46.0\%; removing both yields 36.0\% success and 0.19 unsafe/early termination. These controls support complementary and non-redundant roles: $m_i$ is critical for \emph{whether} to intervene (Rec.\ drops from 0.79 to 0.46 without it), while $z_i$ is critical for \emph{how much} to apply (success drops from 61.3\% to 46.0\% without it). Supplementary multiscale rollouts directly assess the deployed scale choices against observed outcomes.


\paragraph{Effect-guided modeling.}
Table~\ref{tab:egm_offline} isolates the contribution of each EGM design choice by progressively adding components to a direct score-regression baseline. Pairwise ranking replaces pointwise regression; outcome-gap weighting emphasizes informative pairs; and restoration-aware weighting down-weights unreliably restored branches. The cumulative effect improves P$>$B / B$>$N ordering from 0.63 / 0.60 to 0.83 / 0.88 and reduces FPR@80\% from 0.40 to 0.12, translating to a 23.3pp end-to-end success gain (38.0\% $\to$ 61.3\%).

\begin{table}[!ht]
\centering
\begin{tabular*}{\columnwidth}{@{\extracolsep{\fill}}lcccc@{}}
\toprule
Variant & P$>$B $\uparrow$ & B$>$N $\uparrow$ & FPR $\downarrow$ & Avg. (\%) $\uparrow$ \\
\midrule
RawScore & .63 & .60 & .40 & 38.0\% \\
PairRank & .73 & .72 & .25 & 40.0\% \\
+ Gap & .71 & .76 & .29 & 44.0\% \\
+ Restore & \textbf{.83} & \textbf{.88} & \textbf{.12} & \textbf{61.3\%} \\
\bottomrule
\end{tabular*}
\caption{EGM ablation on held-out branch groups. FPR is measured at 80\% recall.}
\label{tab:egm_offline}
\end{table}

\paragraph{Residual action modulation.}
Table~\ref{tab:ram_offline} evaluates how EGM's effect evidence is best converted into a deployment-time action choice. Using EGM scores directly (EGM-Greedy) or restricting supervision to a single evidence source (Outcome-only, Effect-only) each degrades performance. The full joint selector, conditioned on both EGM rankings and branch-outcome evidence, attains the strongest held-out top-1 agreement (0.79), paired recovery (0.79), and average success (61.3\%), indicating that both evidence sources contribute non-redundantly.

\begin{table}[!ht]
\centering
\begin{tabular*}{\columnwidth}{@{\extracolsep{\fill}}lccc@{}}
\toprule
Variant & Off. T1 $\uparrow$ & Rec. $\uparrow$ & Avg. (\%) $\uparrow$ \\
\midrule
EGM-Greedy & .45 & .41 & 36.0\% \\
Outcome-only & .64 & .56 & 42.0\% \\
Effect-only & .51 & .50 & 42.0\% \\
Unweighted & .68 & .61 & 47.0\% \\
Two-stage & .62 & .57 & 44.0\% \\
\textbf{Full RAM} & \textbf{.79} & \textbf{.79} & \textbf{61.3\%} \\
\bottomrule
\end{tabular*}
\caption{RAM ablation. Off. T1 is held-out oracle-positive top-1 agreement; Rec. uses matched-start rollouts.}
\label{tab:ram_offline}
\end{table}

\section{Conclusion}

We presented ViTaR, a framework that reframes tactile feedback from an action-generating perceptual input to a bounded execution modulator for frozen VLA policies. By decomposing contact-aware adaptation into effect-guided residual ranking and tactile-conditioned gain scaling, ViTaR improves the frozen VLA baseline by 30.6 percentage points on the UniVTAC benchmark and demonstrates consistent gains on physical-robot tasks spanning insertion, sliding, and grasping, all without modifying the base policy's pretrained representations. Future work includes extending the paradigm to multi-finger dexterous manipulation with heterogeneous tactile sensors, scaling the branch-comparison protocol to longer-horizon tasks.

\bibliography{aaai2027}

\clearpage
\section*{Supplementary Material}


\section{Additional Ablation Studies}

\subsection{Multiscale Tactile Scaling Evaluation}
\label{sec:multiscale}

\paragraph{Protocol and decision-time inputs.}
The rank-derived auxiliary gain target used during training is not a physical scale label. We therefore directly evaluate the scale predicted at deployment on 240 held-out UniVTAC decision states for which RAM selects a residual direction. The selected direction $u_i^*$ is fixed throughout the comparison, so the experiment isolates scale selection rather than residual-direction selection. From the restored state, we execute the same direction at
\[
\mathcal A=\{0.25,0.50,0.75,1.00\},
\]
with the same intervention horizon, base-policy continuation, action clipping, local outcome protocol, and safety rules. Scale-branch order is randomized within each state group, and each scale is repeated three times. The scale oracle is defined only over this evaluated set,
\[
\alpha_i^{\mathrm{oracle}}
=\arg\max_{\alpha\in\mathcal A}q_i(u_i^*,\alpha).
\]
All learned scale models access observations available before the residual is executed. In particular, the fine tactile summary $z_i$, when present, is computed from decision-time bilateral tactile-image history and contains no post-intervention tactile observation, branch outcome, or future contact measurement.

\paragraph{Controls and metrics.}
\emph{Fixed-1.0} always applies the selected direction at full scale. \emph{No fine tactile} uses the same local state and selected direction as ViTaR but removes $z_i$ from the scale model. \emph{ViTaR scaling} predicts a continuous $\hat\alpha_i$ from $x_i$, $u_i^*$, and $z_i$; for this discrete evaluation, it is projected to the nearest member of $\mathcal A$. The \emph{Discrete oracle} chooses $\alpha_i^{\mathrm{oracle}}$ after observing all four branch outcomes and is therefore an upper reference, not a deployable method.

Exact accuracy measures whether the projected prediction equals $\alpha_i^{\mathrm{oracle}}$, while W1 accuracy allows one adjacent scale bin. Error is the mean absolute distance between the projected prediction and the oracle bin. To calculate local regret, only the four scores within each state-level scale group are rescaled to $[0,1]$, and we compute
\[
q_i(u_i^*,\alpha_i^{\mathrm{oracle}})
-q_i\!\left(u_i^*,\Pi_{\mathcal A}(\hat\alpha_i)\right),
\]
where $\Pi_{\mathcal A}$ is nearest-bin projection. Scores are then aggregated across groups. Branch success and Unsafe. are measured from the rollout at the corresponding scale. These metrics compare the deployed output with observed multiscale outcomes; they do not treat agreement with the rank-derived training target as evidence of a physical optimum.

\paragraph{Results and scope.}
Table~\ref{tab:multiscale_scaling} validates both the scale output and the role of fine tactile information. ViTaR obtains 0.68 exact-scale accuracy and 0.94 W1 accuracy, reducing error to 0.08 and local regret to 0.04. Relative to Fixed-1.0, this reduces regret by 0.11, improves branch success by 13.2 percentage points (48.7\% to 61.9\%), and lowers Unsafe. from 0.12 to 0.04. Removing $z_i$ increases regret to 0.10 and reduces success to 50.8\%, showing that the improvement is not explained by the selected residual direction alone. The discrete oracle reaches 68.4\% success, leaving a 6.5-point gap and thus indicating remaining headroom. The result supports tactile-conditioned selection of a more favorable scale among $\mathcal A$ for an already selected residual direction; it does not claim recovery of a continuous physical optimum outside the evaluated scale set.

\begin{table*}[t]
\centering
\begin{tabular*}{\textwidth}{@{\extracolsep{\fill}}lccccccc@{}}
\toprule
Scale method & States & Exact $\uparrow$ & W1 $\uparrow$ & Error $\downarrow$ & Regret $\downarrow$ & Success $\uparrow$ & Unsafe. $\downarrow$ \\
\midrule
Fixed-1.0                & 240 & .34 & .63 & .27 & .15 & 48.7\% & .12 \\
No fine tactile          & 240 & .46 & .78 & .17 & .10 & 50.8\% & .10 \\
\textbf{ViTaR scaling}  & 240 & \textbf{.68} & \textbf{.94} & \textbf{.08} & \textbf{.04} & \textbf{61.9\%} & \textbf{.04} \\
Discrete oracle          & 240 & 1.00 & 1.00 & .00 & .00 & 68.4\% & \textbf{.03} \\
\bottomrule
\end{tabular*}
\caption{Held-out multiscale evaluation for residual directions selected by RAM. Exact and W1 denote exact-bin and within-one-bin accuracy; Error is mean absolute bin error. Predictions are projected to the nearest evaluated bin; the discrete oracle is not deployable.}
\label{tab:multiscale_scaling}
\end{table*}

\subsection{Restoration Sensitivity and Branch Stability}
\label{sec:sensitivity}

\begin{table*}[t]
\centering
\begin{minipage}[t]{0.48\textwidth}
\centering
\textbf{(a) Restoration-quality treatment}\par
\begin{tabular*}{\linewidth}{@{\extracolsep{\fill}}lcccc@{}}
\toprule
Setting & P$>$B & B$>$N & FPR & Avg. \\
\midrule
Strict-pass & \textbf{.84} & \textbf{.89} & \textbf{.11} & 59.7\% \\
Strict + downwtd. & .83 & .88 & .12 & \textbf{61.3\%} \\
No restoration wt. & .71 & .76 & .29 & 44.0\% \\
Relaxed threshold & .77 & .82 & .20 & 54.8\% \\
\bottomrule
\end{tabular*}
\end{minipage}\hfill
\begin{minipage}[t]{0.48\textwidth}
\centering
\textbf{(b) Repeated branch outcomes}\par
\begin{tabular*}{\linewidth}{@{\extracolsep{\fill}}lccccc@{}}
\toprule
Platform & Groups & $K$ & SD($\Delta q$) & Rank & Labels \\
\midrule
UniVTAC & 245 & 3 & .074 & .83/.72 & .89 \\
Physical & 48 & 3 & .091 & .75/.64 & .84 \\
\bottomrule
\end{tabular*}
\end{minipage}
\caption{Restoration and branch-stability analyses. (a) ``downwtd.'' applies the restoration reliability weight. (b) Rank reports Spearman/Kendall correlation across $K=3$ repeats; the following tables provide complementary restoration and preference-pair analyses.}
\label{tab:sensitivity}
\end{table*}

\paragraph{Restoration-quality treatment.}
Table~\ref{tab:sensitivity}(a) uses the same held-out branch groups for four treatments. \emph{Strict-pass} retains only branches that pass the restoration check. \emph{Strict + downweighted} is the default: it retains the same strict-passing branches and includes the remaining branches with $\kappa=0.25$. \emph{No restoration weight} uses all branches with unit weight, and \emph{Relaxed threshold} uses a less selective restoration check. The comparison changes only the treatment of restoration quality; task stratification, candidate residual options, branch score, EGM architecture, and evaluation protocol are otherwise fixed.

Strict-only filtering gives the strongest offline preference ordering (P$>$B $=0.84$ and B$>$N $=0.89$), but its success is 59.7\%. The default treatment retains nearly the same ordering quality (0.83 and 0.88), attains the lowest high-recall FPR among the retained-data settings (0.12), and gives the strongest end-to-end success, 61.3\%. In contrast, assigning full weight to imperfectly restored branches reduces ordering accuracy to 0.71/0.76, raises FPR to 0.29, and lowers success to 44.0\%. This shows that the reliability factor is consequential rather than cosmetic: it trades a small amount of strict-filtering selectivity for greater usable data while limiting the effect of detected restoration mismatch. It does not establish exact counterfactual equivalence of restored branches.

\paragraph{Downstream consequences of restoration treatment.}
Table~\ref{tab:restoration_sensitivity_supp} reports the complementary selector and matched-start metrics for the same scan. The default strict-plus-downweighted treatment has the highest held-out top-1 agreement (0.79) and paired recovery (0.79). Strict-pass only has the lowest Unsafe. rate (0.03 versus 0.04), consistent with its more selective use of data, whereas no restoration weighting degrades top-1 agreement, recovery, and Unsafe. simultaneously. Together with Table~\ref{tab:sensitivity}(a), these results motivate the default treatment as a practical compromise; they do not turn restoration weighting into a guarantee against unobserved reset differences.

\begin{table}[t]
\centering
\begin{tabular*}{\columnwidth}{@{\extracolsep{\fill}}lccc@{}}
\toprule
Setting & Off. T1 $\uparrow$ & Rec. $\uparrow$ & Unsafe. $\downarrow$ \\
\midrule
Strict-pass only & .78 & .78 & \textbf{.03} \\
Strict + downweighted & \textbf{.79} & \textbf{.79} & .04 \\
No restoration weighting & .63 & .61 & .13 \\
Relaxed threshold & .70 & .70 & .08 \\
\bottomrule
\end{tabular*}
\caption{Supplementary downstream metrics for the restoration-treatment scan in Table~\ref{tab:sensitivity}(a).}
\label{tab:restoration_sensitivity_supp}
\end{table}

\paragraph{Repeated branch outcomes and preference consistency.}
To measure the stability of the local comparison protocol, every base/residual branch in each analyzed group is executed in randomized or balanced order and repeated $K=3$ times. Table~\ref{tab:sensitivity}(b) summarizes 245 task-stratified UniVTAC groups and a 48-group physical subset. SD($\Delta q$) measures variation of the residual--base outcome difference across repeats. Rank reports Spearman/Kendall consistency of residual ordering, and Labels is the agreement of the positive/neutral/negative labels induced by the fixed margin $\delta=0.05$.

The UniVTAC groups exhibit SD($\Delta q$)$=0.074$, rank consistency $0.83/0.72$, and label agreement 0.89. The physical subset is more variable, with SD($\Delta q$)$=0.091$, rank consistency $0.75/0.64$, and label agreement 0.84. Table~\ref{tab:branch_stability_supp} further separates the two pair types used by EGM: positive-over-base agreement is 0.91 in UniVTAC and 0.87 in the physical subset, while base-over-negative agreement is 0.93 and 0.90. These measurements characterize the repeatability of the stated local protocol, including its additional physical variability; they are not evidence that all branches are exact counterfactual interventions.

\begin{table}[t]
\centering
\begin{tabular*}{\columnwidth}{@{\extracolsep{\fill}}lcc@{}}
\toprule
Platform & P$>$B $\uparrow$ & B$>$N $\uparrow$ \\
\midrule
UniVTAC & .91 & .93 \\
Physical subset & .87 & .90 \\
\bottomrule
\end{tabular*}
\caption{Supplementary preference-pair agreement for the repeated-branch analysis in Table~\ref{tab:sensitivity}(b).}
\label{tab:branch_stability_supp}
\end{table}

\subsection{Core Component Ablations}

\paragraph{Effect-guided modeling.}
Table~\ref{tab:supp_egm_offline} uses the same held-out branch groups, state representation, residual options, and evaluation metrics for all EGM rows. \emph{RawScore} replaces local pairwise preference learning with direct regression of a scalar branch score. \emph{PairRank} instead learns the within-group Bradley--Terry ordering. \emph{+ Gap} adds local outcome-gap weighting, and \emph{+ Restore} additionally applies the restoration reliability factor, yielding the full EGM treatment. The comparison tests whether relative local ordering and restoration-aware weighting improve effect evidence beyond direct scalar prediction.

The full treatment improves P$>$B/B$>$N accuracy from 0.63/0.60 for RawScore to 0.83/0.88, reduces FPR at 80\% recall from 0.40 to 0.12, and is associated with 61.3\% end-to-end success versus 38.0\% for RawScore. Because $q_i(c)$ is used only within each branch group, the result supports local preference learning rather than a globally calibrated branch-score interpretation.

\begin{table}[t]
\centering
\begin{tabular*}{\columnwidth}{@{\extracolsep{\fill}}lcccc@{}}
\toprule
Variant & P$>$B $\uparrow$ & B$>$N $\uparrow$ & FPR $\downarrow$ & Avg. (\%) $\uparrow$ \\
\midrule
RawScore & .63 & .60 & .40 & 38.0\% \\
PairRank & .73 & .72 & .25 & 40.0\% \\
+ Gap & .71 & .76 & .29 & 44.0\% \\
+ Restore & \textbf{.83} & \textbf{.88} & \textbf{.12} & \textbf{61.3\%} \\
\bottomrule
\end{tabular*}
\caption{EGM ablation on held-out branch groups. FPR is measured at 80\% recall.}
\label{tab:supp_egm_offline}
\end{table}

\paragraph{Residual action modulation.}
Table~\ref{tab:supp_ram_offline} holds the frozen EGM checkpoint, residual option set, state representation, and deployment action space fixed. \emph{EGM-Greedy} directly takes the highest EGM-ranked choice without a learned joint selector. \emph{Outcome-only} and \emph{Effect-only} restrict the selector to one source of within-group supervisory evidence. \emph{Unweighted} removes the example weighting used in its selector-training control, and \emph{Two-stage} separates base/residual choice from residual-direction prediction. \emph{Full RAM} is the single joint selector over the base action and residual options, conditioned on the current state and EGM's within-state ranking evidence.

Full RAM obtains the highest held-out top-1 agreement (0.79), paired recovery (0.79), and average success (61.3\%). The comparison shows that direct ranking, a single evidence source, or a separated gate/direction design is less effective under the shared local-option protocol. It does not require or assume that an EGM score gap and an outcome-score difference are globally commensurate quantities.

\begin{table}[t]
\centering
\begin{tabular*}{\columnwidth}{@{\extracolsep{\fill}}lccc@{}}
\toprule
Variant & Off. T1 $\uparrow$ & Rec. $\uparrow$ & Avg. (\%) $\uparrow$ \\
\midrule
EGM-Greedy & .45 & .41 & 36.0\% \\
Outcome-only & .64 & .56 & 42.0\% \\
Effect-only & .51 & .50 & 42.0\% \\
Unweighted & .68 & .61 & 47.0\% \\
Two-stage & .62 & .57 & 44.0\% \\
\textbf{Full RAM} & \textbf{.79} & \textbf{.79} & \textbf{61.3\%} \\
\bottomrule
\end{tabular*}
\caption{RAM ablation. Off. T1 is held-out oracle-positive top-1 agreement; Rec. uses matched-start rollouts.}
\label{tab:supp_ram_offline}
\end{table}

\paragraph{Direct residual learning and tactile controls.}
For Table~\ref{tab:supp_rl_comparison}, PPO-option selects from the same discrete base-action/residual interface as ViTaR, whereas SAC-residual learns a bounded continuous residual. Both share the frozen OpenVLA-OFT reference, residual coordinates, eligible decision points, task protocol, and safety limits; neither uses EGM or RAM supervision. ViTaR reaches 61.3\% average success with Unsafe. $=0.04$, compared with 36.0\%/0.25 for PPO-option and 32.0\%/0.31 for SAC-residual. This is a controlled comparison of residual-learning formulations around the same reference policy, not a claim about reinforcement learning in general.

Table~\ref{tab:supp_contact_magnitude} separates the contact descriptor $m_i$ used in residual selection from the fine tactile summary $z_i$ used in residual scaling. ``w/o State'' removes $m_i$, ``w/o Tac.'' removes $z_i$, ``w/o Both'' removes both, and ``Fixed Scale'' replaces the predicted gain with a constant. All retain the base policy and residual option set. Removing $m_i$ lowers paired recovery to 0.46; removing $z_i$ lowers average success to 46.0\%; using a fixed scale reaches 45.0\%; and removing both reaches 36.0\% with Unsafe. $=0.19$. These end-to-end controls motivate the distinct roles of contact-aware selection and tactile-conditioned scaling. The multiscale study in Table~\ref{tab:multiscale_scaling} then evaluates the resulting scale output directly against observed scale outcomes.

\begin{table*}[t]
\centering
\begin{minipage}[t]{0.48\textwidth}
\centering
\textbf{(a) Direct residual learning}\par
\begin{tabular*}{\linewidth}{@{\extracolsep{\fill}}lcccc@{}}
\toprule
Method & Rec. $\uparrow$ & Pres. $\uparrow$ & Unsafe. $\downarrow$ & Avg. (\%) $\uparrow$ \\
\midrule
PPO-option & .55 & .66 & .25 & 36.0\% \\
SAC-residual & .45 & .74 & .31 & 32.0\% \\
\textbf{Full ViTaR} & \textbf{.79} & \textbf{.92} & \textbf{.04} & \textbf{61.3\%} \\
\bottomrule
\end{tabular*}
\end{minipage}\hfill
\begin{minipage}[t]{0.48\textwidth}
\centering
\textbf{(b) Contact and scaling controls}\par
\begin{tabular*}{\linewidth}{@{\extracolsep{\fill}}lccc@{}}
\toprule
Variant & Rec. $\uparrow$ & Unsafe. $\downarrow$ & Avg. $\uparrow$ \\
\midrule
w/o State & .46 & .16 & 41.0\% \\
w/o Tac. & .61 & .13 & 46.0\% \\
w/o Both & .40 & .19 & 36.0\% \\
Fixed Scale & .57 & .11 & 45.0\% \\
\textbf{Full ViTaR} & \textbf{.79} & \textbf{.04} & \textbf{61.3\%} \\
\bottomrule
\end{tabular*}
\end{minipage}
\caption{Direct residual-learning and contact/scaling ablations. Rec. and Pres. are matched-start diagnostics; Unsafe. is the safety/early-termination rate.}
\label{tab:supp_rl_comparison}
\label{tab:supp_contact_magnitude}
\end{table*}

\subsection{Structured Residual Options}

\paragraph{Structured residual options.}
To construct the structured set, we aggregate frozen VLA action chunks over the task's training trajectories and retain only coordinate-aligned translational, rotational, or gripper directions corresponding to their predominant signed motion components. This procedure specifies a small set of plausible local correction directions rather than a hand-authored action trajectory, and it is fixed before collecting branch outcomes or evaluating held-out states. The nominal residual range is set by the shared action-safety bounds; no residual direction is tuned with a task-specific execution magnitude. RAM subsequently predicts the state-dependent gain $\alpha_i$, so the candidate set determines ``where to adjust'' while tactile-conditioned scaling determines ``how much to apply.''

Table~\ref{tab:option_set} controls the geometry of the candidate residual set while holding the learned scoring, joint base/residual selector, tactile scaling model, decision points, and evaluation protocol fixed. The matched random control has the same 7D action representation and the same number of candidate entries as the structured dictionary, but does not encode the task-aligned translational, rotational, or gripper adjustments used by ViTaR. Thus, the comparison tests the residual option set rather than increasing the selector action space or model capacity.

Positive-option coverage is the fraction of held-out local branch groups containing at least one residual whose observed score exceeds the base action by the fixed positive margin. Best branch gain is the largest observed residual-over-base score difference within such a group. Both are offline branch-oracle diagnostics: they describe what the available set contains, not a quantity observed by the deployed policy. The structured dictionary increases coverage from 0.41 to 0.72 and mean best gain from 0.13 to 0.34. It also improves paired recovery from 0.40 to 0.79 and preservation from 0.65 to 0.92.

These differences carry through to end-to-end evaluation: the structured set attains 61.3\% average success with Unsafe. $=0.04$, compared with 31.0\% and 0.31 for the matched random set. The result does not claim that the branch oracle is available at deployment. Rather, it shows that a task-aligned, bounded residual dictionary supplies a more useful local action prior for the same learned selection and scaling procedure.

\begin{table}[t]
\centering
\begin{tabular*}{\columnwidth}{@{\extracolsep{\fill}}lccc@{}}
\toprule
Option set & Pos. cov. $\uparrow$ & Best gain $\uparrow$ & Rec. $\uparrow$ \\
\midrule
\shortstack[l]{Matched Random\\7D Set} & .41 & .13 & .40 \\
\shortstack[l]{Structured 7D\\(\textbf{Full ViTaR})} & \textbf{.72} & \textbf{.34} & \textbf{.79} \\
\midrule
Option set & Pres. $\uparrow$ & Unsafe. $\downarrow$ & Avg. (\%) $\uparrow$ \\
\midrule
\shortstack[l]{Matched Random\\7D Set} & .65 & .31 & 31.0\% \\
\shortstack[l]{Structured 7D\\(\textbf{Full ViTaR})} & \textbf{.92} & \textbf{.04} & \textbf{61.3\%} \\
\bottomrule
\end{tabular*}
\caption{Residual option-set control. Positive-option coverage and best branch gain are offline properties measured from held-out branch outcomes. Recovery and preservation use matched-start rollouts; Unsafe. is the safety/early-termination rate.}
\label{tab:option_set}
\end{table}

\section{Experimental Hardware Setup and Data Collection}

\subsection{Data-Collection System}
\begin{figure}[!t]
\centering
\includegraphics[width=\columnwidth]{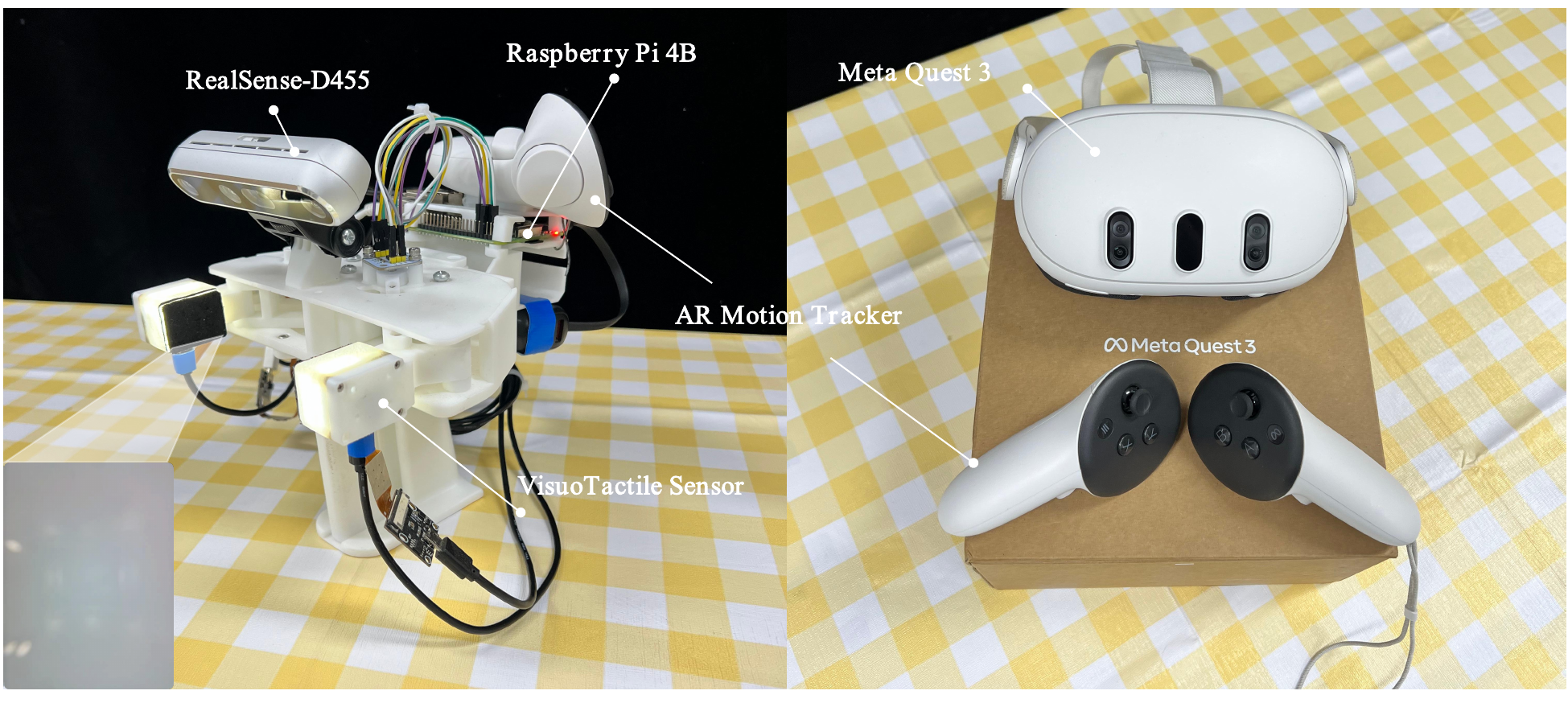}
\caption{Physical demonstration collection with ExUMI and Meta Quest. The teleoperation rig records the operator interface, wrist visual stream, tracked motion, robot proprioception, and bilateral tactile images as synchronized demonstration data.}
\label{fig:data_collection}
\end{figure}

Physical demonstrations are collected with ExUMI, using a Meta Quest headset and its tracked hand controllers for immersive teleoperation (Fig.~\ref{fig:data_collection}). Three operators collect demonstrations. Each task trajectory takes approximately 30--50 seconds, including the approach, contact-rich manipulation, and completion phases. ExUMI records synchronized RGB, robot proprioception, controller pose, and bilateral tactile streams. The headset provides the operator with visual feedback; the controller's 6D pose is mapped, through the standard ExUMI clutching and frame-alignment procedure, to Cartesian end-effector increments, while the gripper command is mapped to the controller's grasp input. This relative mapping avoids discontinuities when the operator repositions the controller and preserves the robot workspace and joint limits.

Before each collection session, we follow the ExUMI calibration procedure: the robot is moved to a repeatable home configuration; the tracker/controller frame is aligned with the robot base frame; the wrist camera is checked after mounting; and the gripper open/close directions are verified. Time-stamped sensor streams are then recorded under the same clocking protocol used by ExUMI. Each 9DTact sensor additionally receives a no-contact reference image before collection. The three operators use the same calibration and reset procedure; operator identity is not used as a policy input.

\subsection{Physical-Robot Hardware}
\begin{figure}[!t]
\centering
\includegraphics[width=0.92\columnwidth]{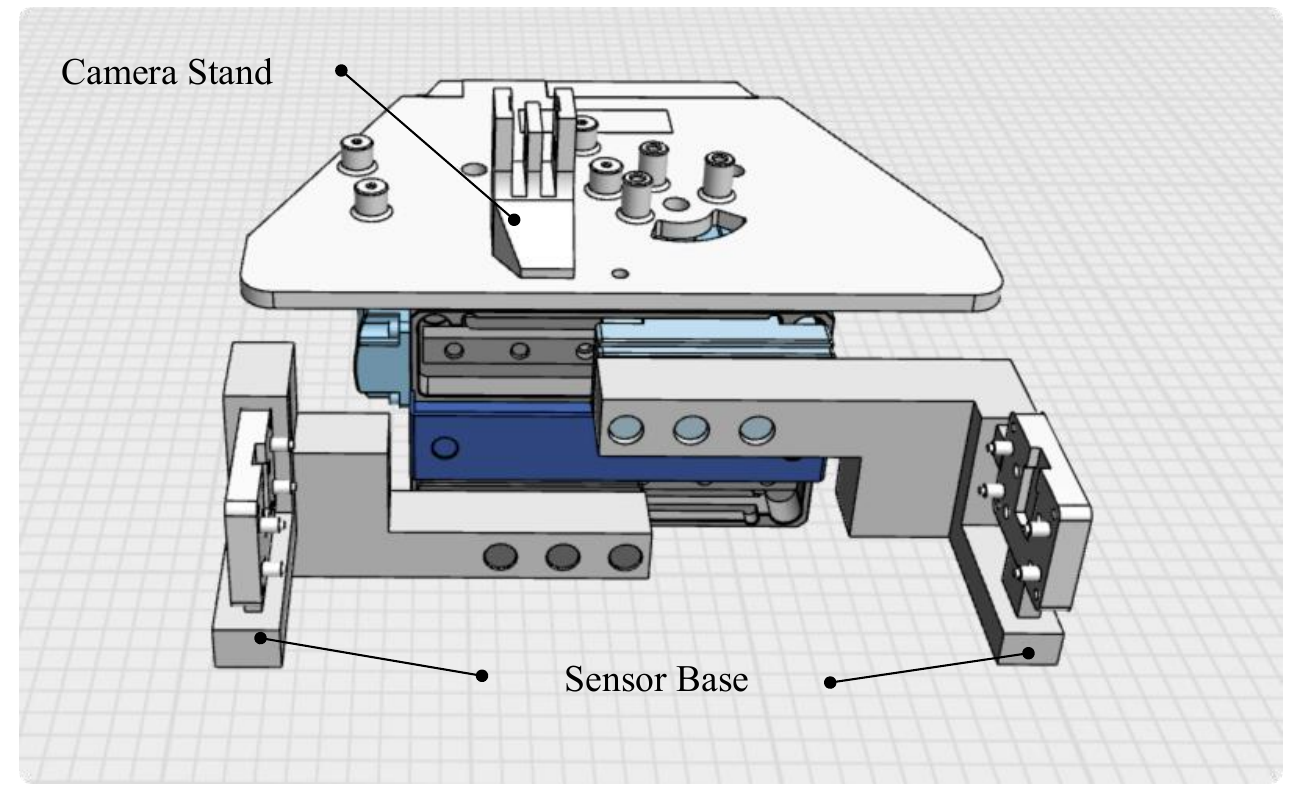}
\caption{Custom 3D-printed adapter for mounting the 9DTact sensing units on the DH Robotics gripper. The drawing shows the sensor-base and camera-stand mounting geometry.}
\label{fig:gripper_adapter}
\end{figure}

The physical platform uses a 6-DoF RealMan RM65-B manipulator (RM-65 series) and a DH Robotics PGIA-series parallel-jaw gripper. We designed and 3D-printed a rigid adapter that fixes two identical 9DTact sensors to the two gripper fingers with mirror-symmetric sensing surfaces. The adapter provides repeatable sensor placement, protects cable routing during opening and closing, and keeps the tactile contact faces inside the gripper workspace. Figure~\ref{fig:gripper_adapter} shows the mounting geometry of the 3D-printed part; it is a hardware fixture rather than a learned component of ViTaR.

\begin{figure}[!t]
\centering
\includegraphics[width=0.90\columnwidth]{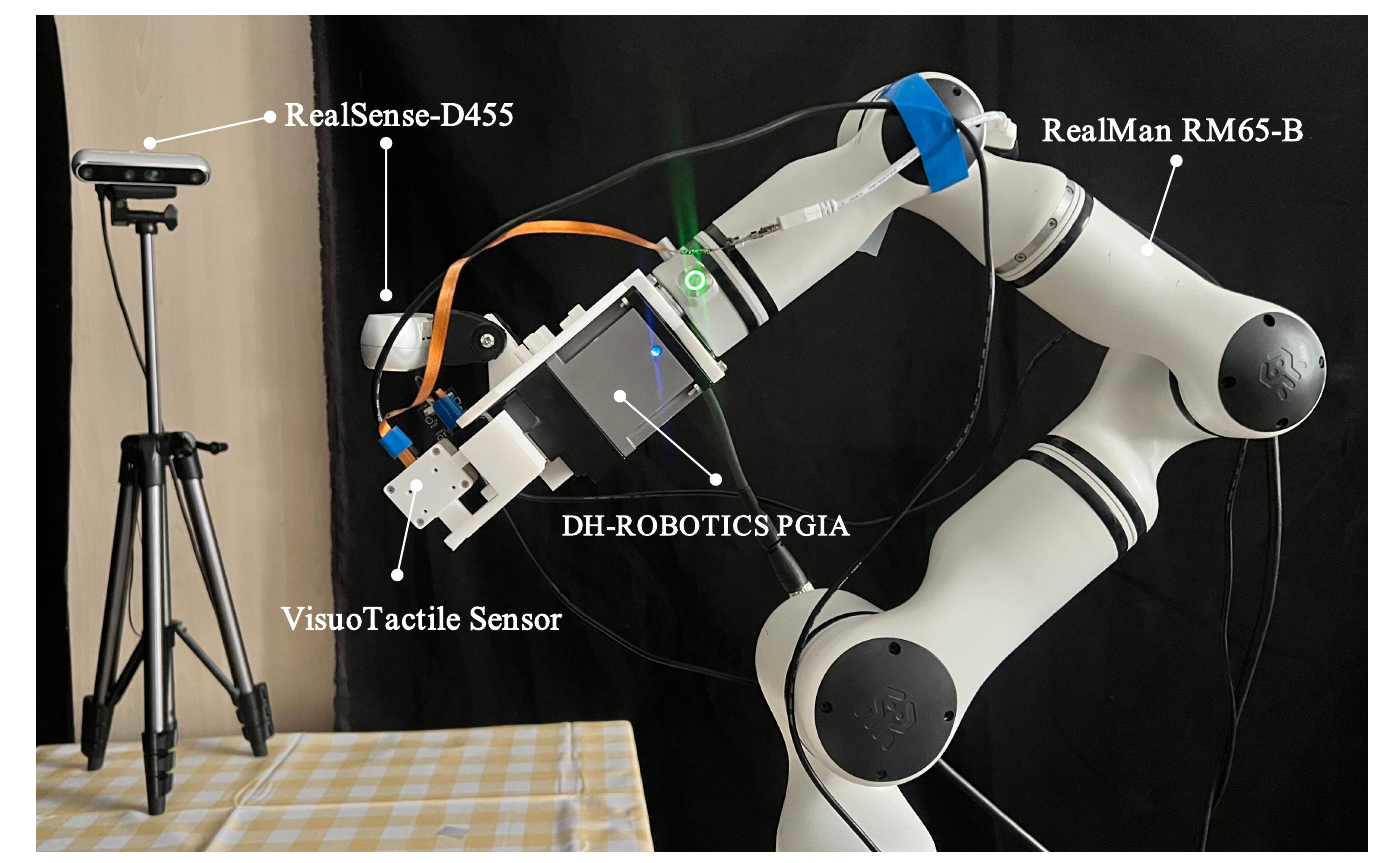}
\caption{Physical-robot hardware used for ViTaR evaluation. The 6-DoF RealMan RM65-B manipulator (RM-65 series) carries the DH Robotics gripper with the custom 3D-printed dual-9DTact adapter. A wrist-mounted RealSense D455 and a second D455 provide ego-centric and third-person views, respectively.}
\label{fig:physical_platform}
\end{figure}

The second hardware view in Fig.~\ref{fig:physical_platform} shows the complete system: the RM65-B (RM-65 series), the DH gripper and its two tactile sensors, a wrist-mounted RealSense D455, and a fixed third-person D455. The wrist camera provides an ego-centric view of the object and contact region, while the third-person camera provides scene context and supports reset verification. The same robot configuration is used for \emph{Insert Hole}, \emph{Lift Bottle}, and \emph{Wiping the Board}. \emph{Lift Bottle} shares contact requirements with its simulated counterpart. \emph{Insert Hole} follows the contact-guided insertion regime represented in the UniVTAC task family, while \emph{Wiping the Board} additionally requires sustained sliding contact.

\subsection{Sensing, Calibration, and Control Timing}
Each 9DTact unit is operated at $640\times480$ pixels and 30 Hz, following the public reference configuration accompanying the 9DTact paper. The focus is adjusted so that the sensing surface at approximately 15 mm is sharp. For each sensor, calibration starts with an averaged no-contact reference image. A $7\times9$ calibration grid with 2.5 mm spacing is used to rectify image coordinates and determine the pixel scale. We then use a 4 mm-radius spherical indenter to construct the sensor-specific intensity-to-depth lookup used by the released 9DTact calibration procedure. During operation, raw tactile frames are rectified, cropped to the calibrated sensing region, and paired with the corresponding no-contact reference before the tactile summary is computed. This procedure preserves the bilateral geometry while avoiding an assumption that the two sensor optical systems are identical.

Both D455 color cameras are configured at $1280\times720$ pixels and 30 Hz. We use fixed manual exposure after an initial scene-specific adjustment (8 ms under the indoor lighting used for collection) to prevent exposure changes from becoming a spurious visual cue; images are resized only at the model input stage. Rather than sharing nominal intrinsics across devices, we retain the factory RGB intrinsics for each D455 and verify them with a checkerboard calibration after mounting. The wrist-camera-to-end-effector transform is obtained with standard hand--eye calibration, and the third-person-camera-to-robot-base transform is obtained from a checkerboard/AprilTag board fixed in the workspace. These per-device intrinsic matrices and rigid transforms are applied to the recorded data, so the two camera views remain geometrically consistent even though their mount positions differ.

The physical controller runs at 10 Hz, a common VLA manipulation rate that provides a 100 ms action period while retaining stable contact execution. Camera and tactile streams are captured at 30 Hz; the latest synchronized observations are selected at each 10 Hz decision instant. The frozen VLA emits an action chunk, and the residual module either preserves it or applies one bounded 7D correction to every action in the chunk. Commands are clipped to the pre-specified action and hardware-safety bounds before dispatch. Early termination is triggered by the same safety checks during both branch collection and deployment, and the frozen base policy is never updated online.

\subsection{Compute and Deployment Host}
Training and offline processing use a server with eight NVIDIA RTX 4090 GPUs and an Intel Xeon(R) Gold 6430 CPU, running Ubuntu 22.04 LTS with NVIDIA driver version 580.105.08. The GPU configuration is used for foundation-policy fine-tuning, EGM/RAM optimization, and offline branch-data processing; deployment uses the same software stack with the frozen policy and residual modules in inference mode. The robot-facing service is scheduled to the 10 Hz control cycle. Thus, visual/tactile synchronization, inference, residual selection, and command dispatch are budgeted within one 100 ms control period; the system sends the command associated with the latest complete synchronized observation set rather than blocking the robot for a late sensor frame.

\subsection{Physical Evaluation Protocol}
Each physical method--task pair is evaluated over 20 end-to-end trials under the common hardware configuration. During both branch collection and deployment, residual exploration is restricted to the pre-specified 7D option set, action bounds, hardware safety limits, and early-termination rules. Thus, physical branches are controlled local comparisons rather than open-ended policy exploration; this protocol limits the scope of the safety claim to the evaluated candidate set and tasks.

\section{Task Details}

\subsection{Simulated UniVTAC Tasks}
\begin{figure*}[!t]
\centering
\includegraphics[width=0.72\textwidth]{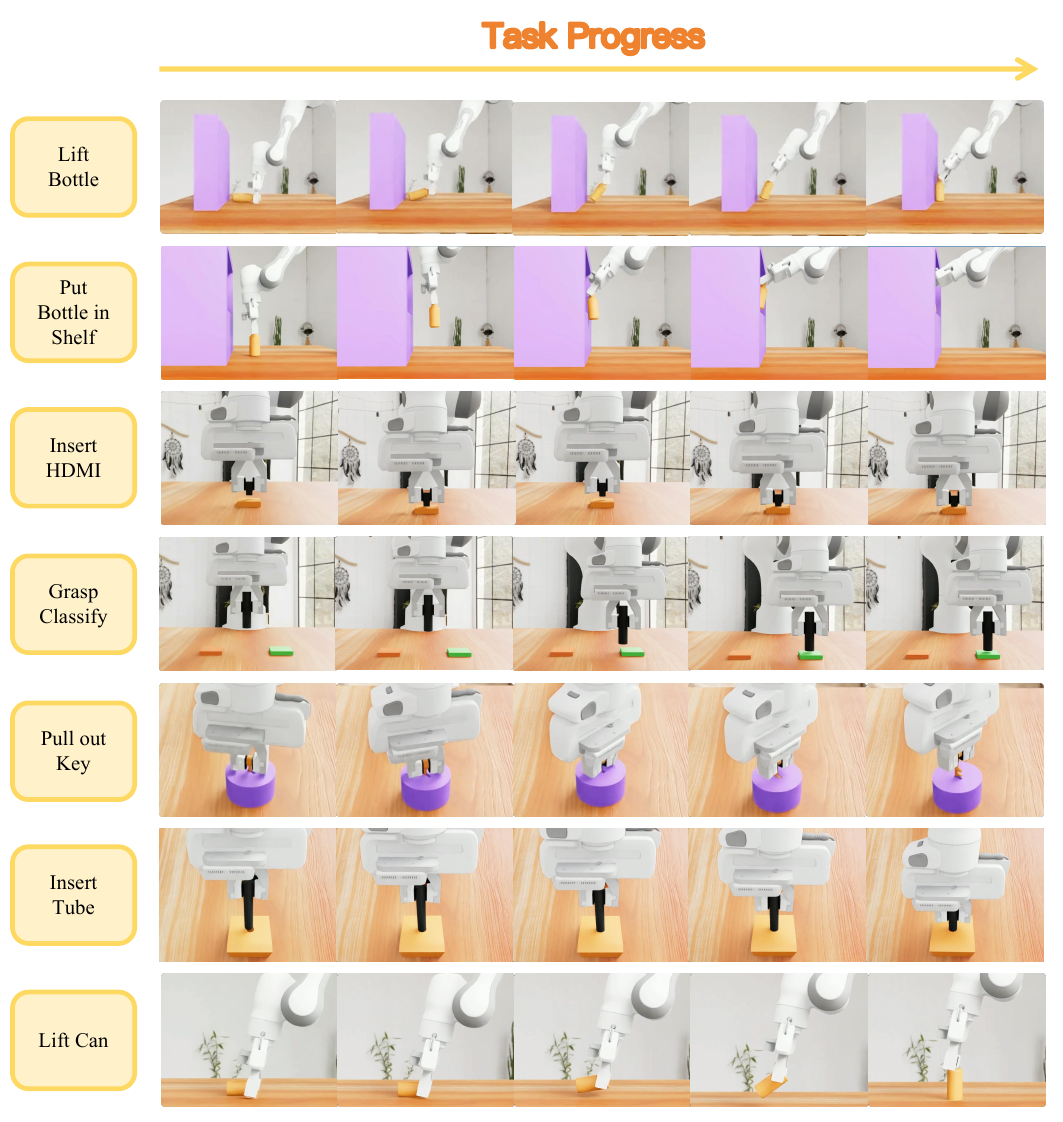}
\caption{Complete workflows for the seven simulated UniVTAC tasks. Each row shows the initialized scene and its task-progress sequence through the terminal goal state.}
\label{fig:univtac_task_workflows}
\end{figure*}

The simulated suite contains \emph{Lift Bottle}, \emph{Pull-out Key}, \emph{Lift Can}, \emph{Put Bottle}, \emph{Insert HDMI}, \emph{Insert Tube}, and \emph{Grasp Classify}. Figure~\ref{fig:univtac_task_workflows} summarizes their start-to-goal sequences. Together, these tasks cover grasping, pulling, constrained placement, insertion, and tactile discrimination with synchronized visual, proprioceptive, and bilateral tactile observations.

\paragraph{Common protocol.}
An episode succeeds when its task-side goal is reached while contact remains physically valid. The UniVTAC checker ends a rollout protectively for excessive contact penetration or significant gelpad--object slip; a valid episode that has not reached the goal at its horizon is incomplete.

\paragraph{Lift Bottle.}
The robot grasps a bottle resting near a wall and lifts it vertically. Success requires the lifted bottle base to finish within 5~cm of the wall, testing stable grasping and relative-pose control.

\paragraph{Pull-out Key.}
Starting from a randomly rotated key in a slot, the robot grasps, rotates the key to the compatible orientation using resistance cues, and pulls it along the slot axis. Success is removal of the key.

\paragraph{Lift Can.}
The robot grasps and vertically lifts a horizontal cylindrical can of 4, 5, or 6~cm diameter. Success requires a slip-free lift, probing contact-conditioned grasp aperture across the three geometries.

\paragraph{Put Bottle.}
The robot grasps an upright bottle, transports it to a shelf, aligns it with the opening, and places it in the shelf cavity. Success is occupancy of the cavity.

\paragraph{Insert HDMI.}
The robot refines the alignment of an HDMI connector with randomized rotational offset and inserts it into a fixed slot. Success is a seated connector, making this a compact precision-insertion task.

\paragraph{Insert Tube.}
The robot aligns a 2.0~cm test tube with a 2.05~cm hole on an inclined surface and advances it through the opening. Success is completed insertion through this tight clearance.

\paragraph{Grasp Classify.}
The robot uses tactile contact to distinguish two visually similar cylinders by texture and places the selected object in its class-specific goal region. Success requires both correct classification and placement.

\subsection{Physical-Robot Tasks}
\begin{figure*}[!t]
\centering
\includegraphics[width=0.90\textwidth]{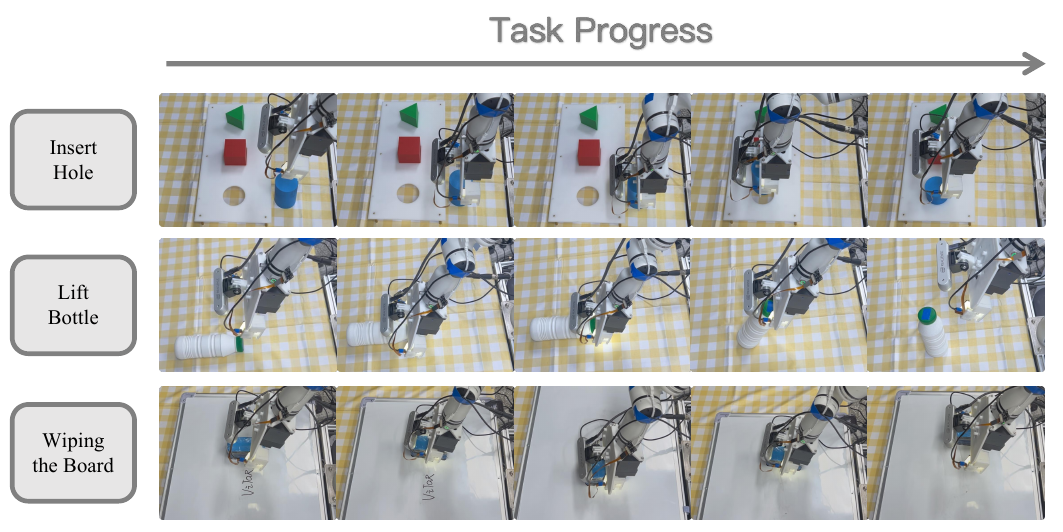}
\caption{Complete physical-robot task workflows for \emph{Insert Hole}, \emph{Lift Bottle}, and \emph{Wiping the Board}.}
\label{fig:physical_task_workflows}
\end{figure*}

The physical suite uses the RM65-B/DH-gripper platform for \emph{Insert Hole}, \emph{Lift Bottle}, and \emph{Wiping the Board}; Fig.~\ref{fig:physical_task_workflows} shows their task-progress sequences. Each method--task pair is evaluated in 20 trials. A trial succeeds at its visible goal state; hardware-safety stops and horizon-limited trials are reset as not completed.

\paragraph{Insert Hole.}
The robot uses light contact to align the task object with the target aperture and advances it to the seated state. Success is completed insertion.

\paragraph{Lift Bottle.}
The robot establishes bilateral tactile contact, closes the gripper, and lifts and retains the bottle. Success is stable retention at the terminal lift state.

\paragraph{Wiping the Board.}
The robot establishes gentle contact with the board and completes the designated wiping sweep while maintaining sliding contact. Success is completion of the intended wiping pass.

\section{Model Training, Inference, and Evaluation Details}

\subsection{Shared Experimental Conventions}

\paragraph{Local branch groups.}
All branch-based analyses use a local decision group containing a base-action branch and the residual branches available at the same, or sufficiently similar, restored decision point. The base branch retains the frozen VLA action chunk. A residual branch applies one fixed 7D option to every action step in that chunk, follows the same intervention horizon and action clipping, and then resumes the base-policy continuation. The residual option set, base policy, local score definition, and safety limits are held fixed when comparing EGM, RAM, restoration, and option-set variants.

\paragraph{Outcome and restoration convention.}
The local branch score combines task completion and progress with failure or early-stop penalties, marker-derived contact retention, and a restoration-quality term. It is used only to compare choices within a decision group. A restoration check examines the recovered contact and state condition before a branch is launched. In the default treatment, branches that fail this check are retained but receive the reliability factor $\kappa=0.25$ in the pairwise ranking loss; branches that pass, and the base branch, receive factor one. This reduces the influence of detectable reset mismatch without interpreting approximate restoration as an exact counterfactual. The restoration analyses below explicitly vary this treatment rather than assuming that the weighting eliminates all history or deformation differences.

\paragraph{Metric convention.}
End-to-end success is the principal deployment metric. For matched-start analyses, Rec. is the fraction of cases in which a method succeeds when the base branch fails, and Pres. is the fraction in which both the method and base branch succeed. Unsafe. is the fraction of branches that terminate early or trigger a safety constraint. P$>$B and B$>$N denote held-out preference-ordering accuracy for positive-over-base and base-over-negative pairs; Off. T1 measures agreement with the best observed positive residual in a held-out group. These quantities are offline or post-hoc diagnostics and are never inputs to the deployed policy.

\paragraph{Shared matched-start cohort and denominators.}
All Full ViTaR/Full RAM rows that report matched-start diagnostics in Tables~\ref{tab:supp_rl_comparison}, \ref{tab:supp_contact_magnitude}, \ref{tab:supp_ram_offline}, and \ref{tab:option_set} use the same fixed matched-start cohort $\mathcal M$, the same frozen base-policy rollouts, and the same execution and safety protocol. The full configuration is evaluated once on this cohort; its diagnostic values are then reported consistently across the ablation tables as Rec.$=0.79$, Pres.$=0.92$, and Unsafe.$=0.04$. Let $\mathcal M_{\mathrm{fail}}\subset\mathcal M$ and $\mathcal M_{\mathrm{succ}}\subset\mathcal M$ denote starts on which the paired base rollout fails and succeeds, respectively. Let $S_j$ and $U_j$ indicate ViTaR success and an unsafe/early-termination event at start $j$. The denominators are
\[
\begin{aligned}
\mathrm{Rec.}&=\frac{\sum_{j\in\mathcal M_{\mathrm{fail}}}S_j}{|\mathcal M_{\mathrm{fail}}|},
&\mathrm{Pres.}&=\frac{\sum_{j\in\mathcal M_{\mathrm{succ}}}S_j}{|\mathcal M_{\mathrm{succ}}|},\\
\mathrm{Unsafe.}&=\frac{\sum_{j\in\mathcal M}U_j}{|\mathcal M|}.&&
\end{aligned}
\]
Thus Rec. and Pres. condition on complementary base-outcome subsets of the same start pool, whereas Unsafe. uses all starts in that pool. The multiscale and restoration-stability analyses are explicitly labeled as separate diagnostic cohorts and are not alternate denominators for these canonical full-model values.

\subsection{Training Data and Local Branch Supervision}

\paragraph{Additional branch supervision.}
ViTaR uses the 50 OFT trajectories per UniVTAC task to establish the frozen OpenVLA-OFT reference, and additionally collects local branch rollouts to train EGM and RAM. The branch rollouts supply the task-progress, contact, restoration, and outcome measurements used for local preference labels; they are therefore additional supervision, not a relabeling of the 50 demonstrations. Crucially, the 100 rollouts per task are not 100 independent task demonstrations: they comprise five short branches from each of 20 restored decision groups ($20<50$), whose source states are drawn from the same 50-trajectory demonstration corpus. The 160 preference pairs reuse outcomes within these 20 groups and introduce no additional trajectories. Thus, ViTaR does not use a larger underlying demonstration corpus or more independent long-horizon demonstration trajectories than the 50-trajectory OFT setting; the branch counts instead measure local comparison supervision. Table~\ref{tab:data_budget} reports the available counting units separately. A candidate rollout and a preference pair are not interchangeable: one rollout can participate in multiple pairs. Likewise, the collection, repeated-stability, multiscale, and main-evaluation cohorts serve different analyses and must not be summed into a single ``total episode'' budget.

\begin{table*}[t]
\centering
\begin{tabular*}{\textwidth}{@{}p{0.22\textwidth}p{0.31\textwidth}p{0.41\textwidth}@{}}
\toprule
Record & Reported amount & Role \\
\midrule
OFT demonstrations & 50 trajectories/task; 350 total & Fine-tune the frozen OpenVLA-OFT reference \\
Offline branch collection & 20 restored groups/task from the 50-trajectory corpus; 5 short branches/group = 100 rollouts and 160 derived pairs/task; 140/700/1120 total & Construct EGM/RAM local supervision \\
Repeated-branch analysis & 245 UniVTAC groups and 48 physical groups; $K=3$ & Measure outcome, rank, and label stability \\
Multiscale scale analysis & 240 states $\times$ 4 scales $\times$ 3 repeats $=2{,}880$ executions & Compare deployed scales with observed outcomes \\
Main evaluation & 100 UniVTAC episodes/task; 20 physical trials/task & Report end-to-end task success \\
\bottomrule
\end{tabular*}
\caption{Reported data and evaluation counts. Counts are given in their native units because branch groups, candidate rollouts, preference pairs, and end-to-end evaluations answer different questions.}
\label{tab:data_budget}
\end{table*}

The reported branch budget is a rollout-level accounting rather than an environment-step budget. For the simulated collection script, a branch applies the selected residual for one VLA action chunk, follows the frozen reference policy for up to 30 continuation chunks, and is capped at 70 chunks in total. Early termination can shorten an individual rollout, and an action chunk contains multiple low-level control actions; we therefore do not convert the above counts into a single number of environment interactions. Similarly, the physical stability cohort reports 48 branch groups with three repeats, but the manuscript does not claim an aggregate count of physical residual interventions or a standardized data-collection time because candidate counts and rollout durations vary by task. These unreported quantities should not be inferred from the branch-group counts.

\paragraph{Baseline provenance and comparison scope.}
Table~\ref{tab:baseline_provenance} distinguishes imported benchmark values from author-evaluated baselines. All comparisons use the named task definitions and report the same end-to-end success metric; author-evaluated UniVTAC methods use the seven-task, 100-episode-per-task protocol, and physical methods use 20 trials per task. However, a common task interface is weaker than an equal-data or exactly seed-matched comparison. In particular, ViTaR receives additional branch supervision, while public baseline results may originate from different training pipelines, checkpoints, or random-seed manifests. We therefore interpret the main table as a performance comparison under a common task suite, not as a controlled comparison of equal training data, collection cost, or total environment interaction.

\begin{table*}[t]
\centering
\begin{tabular*}{\textwidth}{@{}p{0.21\textwidth}p{0.25\textwidth}p{0.48\textwidth}@{}}
\toprule
Method group & Result source & Scope and limitation \\
\midrule
ACT, VITaL, ACT+UniVTAC & UniVTAC benchmark report & Imported task-level values; not re-trained or re-evaluated by us \\
OpenVLA-OFT and ViTaR & Author-evaluated & Shared frozen-reference setting; ViTaR additionally uses local branch supervision \\
Tactile-VLA & Author-adapted and evaluated & Adapted to the UniVTAC observation/control interface; upstream training data remain method specific \\
$\pi_{0.5}$ & Author-evaluated open-source checkpoint & Same seven-task evaluation protocol; upstream pretraining and checkpoint provenance are external \\
PPO-option, SAC-residual & Author-evaluated controlled ablations & Same reference, residual interface, decision points, and safety limits; not a matched interaction-budget study \\
\bottomrule
\end{tabular*}
\caption{Baseline provenance and comparison scope. ``Author-evaluated'' means evaluated by us under the stated task protocol; it does not imply that all upstream pretraining or training data are identical across methods.}
\label{tab:baseline_provenance}
\end{table*}

For imported UniVTAC rows, the original training data, initial-state sampling, and evaluation seeds are controlled by the benchmark report rather than this work. For author-evaluated rows, we use the same task definitions and nominal evaluation episode counts, but do not claim a per-episode seed match with imported benchmark values or identical upstream pretraining data across foundation models. Matched-start statements are reserved for the explicitly paired Rec. and Pres. analyses, not for the full main-baseline table. This distinction prevents the reported success differences from being read as evidence of equal data efficiency or matched-seed statistical superiority.

\subsection{Reported Training and Inference Settings}
\label{sec:reported_settings}

This section reports the implementation-level settings used for EGM and RAM. They instantiate, but do not alter, the main-paper method: EGM is trained first and frozen; RAM then uses the frozen EGM evidence together with decision-time observations. All branch outcomes, local scores, and preference labels are used only during offline target construction and are unavailable to both models at deployment.

\subsubsection{Effect-Guided Modeling Architecture and Optimization}

EGM receives the current local state $x_i$, which concatenates visual--proprioceptive observations, language instruction, execution history, the marker-derived contact descriptor, and the frozen base-action encoding. Its state encoder consists of two linear layers with ReLU activations, and its candidate encoder consists of one linear layer followed by ReLU. The effect head concatenates the encoded state and candidate option, then applies a linear--ReLU--linear MLP to produce the scalar effect score. A separate linear--ReLU--linear gate head produces the adaptability logit. All EGM hidden layers have width 128.

We optimize EGM with AdamW for 120 epochs, using learning rate $10^{-3}$, weight decay $10^{-4}$, batch size 128, and seed 0. The ranking and adaptability losses have equal relative weight ($\lambda_{\mathrm{adapt}}=1.0$). Positive, neutral, and negative residual labels use $\delta=0.05$; a branch that fails the restoration check receives the reliability factor $\kappa=0.25$ in the ranking loss. After training, the EGM checkpoint is frozen before RAM targets are constructed.

\subsubsection{Residual Action Modulation Architecture and Optimization}

RAM contains a discrete selector and a continuous magnitude model. The selector is a three-linear-layer MLP with two width-128 ReLU hidden layers and a final output layer whose dimension equals the current number of action classes (the reference continuation plus the available residual options). It predicts a top-1 action choice from the current state, option set, EGM adaptability evidence, and EGM's within-group ordinal ranking evidence. The magnitude model uses the same three-linear-layer, width-128 MLP pattern. Its input is $x_i$, the selected residual option, and the bilateral tactile summary $z_i$; it outputs one gain logit. During training, the magnitude loss is computed with \texttt{BCEWithLogits} on that logit. At deployment, a single sigmoid maps the logit to $\alpha_i\in[0,1]$, avoiding a second sigmoid in the optimization path.

The selector uses cross-entropy with a lightweight residual-mass regularizer; the magnitude model uses the weighted binary cross-entropy described above. RAM uses the same AdamW configuration as EGM (learning rate $10^{-3}$, weight decay $10^{-4}$, batch size 128, 120 epochs, and seed 0), with $\lambda_\alpha=1.0$. The implementation keeps the following RAM-target constants fixed: $\rho=0.50$, $\tau_A=0.15$, $\tau_g=0.20$, and $c_\alpha=0.50$. These are target-construction constants only: they do not expose a reward, branch outcome, or future observation to the deployed policy, and they do not introduce an additional deployment-time rejection gate. The tactile feature $z_i$ is a lightweight bilateral image-summary vector rather than the output of a separately trained tactile encoder; it modulates residual magnitude only and cannot change the selected residual direction or candidate set.

\subsubsection{Simulated Local-Branch Collection}

Each training group starts from a restored local decision point and loads the frozen reference VLA action chunk. We execute the reference continuation and each available structured residual option from that local state. A residual branch modifies one action chunk (\texttt{intervention\_chunk\_count}$=1$), then resumes the reference policy for up to 30 chunks (\texttt{continuation\_chunks}$=30$); the entire local rollout is capped at 70 chunks (\texttt{max\_chunks}$=70$). At rollout completion, the collector records task outcome, local progress, contact retention, marker motion, and restoration quality. These post-rollout quantities create positive, neutral, and negative residual labels only within the corresponding branch group, providing EGM ranking supervision and RAM selection targets. They are never part of the inference input.

\subsubsection{Deployment Inputs and Action Interface}

At deployment, the frozen OpenVLA-OFT reference policy receives the language instruction, primary scene/head RGB image, wrist RGB image, and proprioception. ViTaR additionally receives the current marker/contact descriptor and the bilateral tactile summary. The RAM selector chooses the top-1 element of $\{\text{reference continuation}\}\cup\mathcal R_i$. If it chooses the reference continuation, no residual is applied. Otherwise, the magnitude model predicts $\alpha_i$ and executes $A_{\mathrm{exec}}=A_{\mathrm{ref}}+\alpha_i d(u_i^*)$. The implementation executes one open-loop step per decision (\texttt{num\_open\_loop\_steps}$=1$) with delta end-effector actions. The gripper command is absolute, represented in percentage units, and has maximum position $q_{\max}=0.039$.

\paragraph{Deployment settings.}
The deployed controller uses the frozen OpenVLA-OFT reference together with a fixed, task-aligned 7D residual dictionary. When RAM selects a residual, the same residual is applied uniformly to every action in the selected VLA chunk. The visual inputs are the primary scene/head camera and wrist camera, and image preprocessing uses a center crop with BGR channel order; proprioception is provided alongside these visual observations. Actions use the \texttt{delta\_ee} representation with action scale one for all seven dimensions. The gripper uses an absolute percentage command with $q_{\max}=0.039$.

Before dispatch, we clip the position, rotation, and gripper commands per action by $(0.03, 0.2, 0.02)$, respectively, in the corresponding controller command units. Each decision uses one open-loop step, and $\alpha_i\in[0,1]$ scales only an already selected residual direction. For the held-out multiscale analysis, the deployed prediction is projected to $\{0.25,0.50,0.75,1.00\}$. Training labels, local scores, rewards, future branch outcomes, and re-enumerated branch rollouts are unavailable at deployment. We report 100 end-to-end episodes for each simulated method--task pair and 20 end-to-end trials for each physical method--task pair; all learned decisions use only the current decision-time observation set.


\end{document}